\documentclass[journal]{IEEEtran}%
 \pdfoutput=1
\ifCLASSINFOpdf%
\else%
\fi%
\usepackage{cite}%
\usepackage{amsmath,amssymb,amsfonts}%
\usepackage{algorithm}%
\usepackage{algorithmic}%
\usepackage{graphicx}%
\usepackage{textcomp}%
\usepackage{mathrsfs}%
\usepackage{subfigure}%
\usepackage{multirow,booktabs,color,soul,threeparttable}%
\usepackage[justification=centering]{caption}%
\usepackage{lineno}
\definecolor{hl}{rgb}{0.75,0.75,0.75}%
\sethlcolor{hl}%
\protected\gdef\tw#1{%
\let\oldfootnote\footnote%
{\color{red}{%
\def\footnote##1{\oldfootnote{\color{red}{##1}}}%
#1}}%
\let\footnote\oldfootnote%
}%
\begin{document}%
\title{A New Many-Objective Evolutionary Algorithm Based on Determinantal Point Processes}%
\author{Peng~Zhang,
        Jinlong~Li,~\IEEEmembership{Member,~IEEE,}
        Tengfei~Li,
        and~Huanhuan Chen,~\IEEEmembership{Senior~Member,~IEEE}
\thanks{P. Zhang, J. Li, T. Li and H. Chen  are with School of Computer Science and
  Technology, University of Science and Technology of China, Hefei 230027,
  China (e-mail: zp1996@mail.ustc.edu.cn; jlli@ustc.edu.cn; ltf91@mail.ustc.edu.cn; hchen@ustc.edu.cn).  }
\thanks{Corresponding author: Jinlong Li}
\thanks{This research work was supported by the National Natural Science Foundation of China under Grants 61573328}}
%
%
%
\maketitle%
\begin{abstract}%
To handle different types of Many-Objective Optimization Problems (MaOPs), Many-Objective Evolutionary Algorithms (MaOEAs) need to simultaneously maintain convergence and population diversity in the high-dimensional objective space.
In order to balance the relationship between diversity and convergence, we introduce a Kernel Matrix and probability model called Determinantal Point Processes (DPPs).
Our Many-Objective Evolutionary Algorithm with Determinantal Point Processes (MaOEADPPs) is presented and compared with several state-of-the-art algorithms on various types of MaOPs \textcolor{blue}{with different numbers of objectives}.
The experimental results demonstrate that MaOEADPPs is competitive.
\end{abstract}%
\IEEEpeerreviewmaketitle%
\section{INTRODUCTION}%
\IEEEPARstart{M}{any} practical problems can be treated as Many-objective Optimization Problems (MaOPs), such as ensemble learning~\cite{5416712,Chen2007Evolutionary}, engineering design~\cite{Fleming2005Many}, time series learning ~\cite{2018Multiobjective}. An MaOP can be expressed by Equation~(\ref{1234}):%
\begin{equation}
\begin{split}
{ \rm M}&{\rm inimize}\quad \boldsymbol{F(x)}=(f_{1}(\boldsymbol{x}),f_{2}(\boldsymbol{x}),\dots,f_{M}(\boldsymbol{x}))^{T} \\
{ \rm su}&{\rm bject\:to}\quad \boldsymbol{x} \in \boldsymbol{\Omega},
\end{split}
\label{1234}
\end{equation}%

\noindent where $\boldsymbol{x} = (x_{1}, x_{2},\dots, x_{n})^{T}$ is an $n$-dimensional decision vector in the decision space~$\boldsymbol{\Omega}$, $M>3$~is the number of objectives, and $\boldsymbol{F}(x)$~is the $M$-dimensional objective vector.

\textcolor{blue}{When solving MaOPs with Many-Objective Evolutionary Algorithms (MaOEAs), the incomparability of solutions causes the proportion of nondominated solutions to increase enormously~\cite{li2015many,Fonseca1998Multiobjective}.} This makes optimization using only the domination relationship infeasible. It is also difficult to maintain population diversity in a high-dimensional objective space~\cite{li2015many}. In order to solve these two problems, researchers have proposed \textcolor{blue}{several methods}, most of which can be roughly divided into three categories.

The Pareto dominance-based MaOEAs belong to the first category, which employs modified Pareto dominance-based mechanisms to select nondominated solutions. For instance, \mbox{$\epsilon$-dominance~\cite{hadka2013borg}} and fuzzy dominance~\cite{he2014fuzzy} employ modified definitions of dominance to maintain the selection pressure. \textcolor{blue}{Different distance measures are used to improve the performance of Pareto dominance-based MaOEAs~\cite{Tanabe2017Benchmarking,8543664}.} Zhang \emph{et~al.} employ a knee point-based selection scheme to select nondominated solutions~\cite{zhang2015knee}. Li \emph{et~al.} propose a shift-based density estimation strategy to make Pareto dominance-based algorithms more suitable for many-objective optimization~\cite{li2014shift}.

Indicator-based MaOEAs form the second category. These MaOEAs use indicators to evaluate the solutions and guide the search process. The hypervolume (HV)~\cite{while2006faster,Emmerich2005An} is one \textcolor{blue}{widely used indicator. The HypE employs Monte Carlo simulation to estimate the HV contribution of the candidate solutions and reduce the high complexity of calculating the HV \cite{bader2011hype}.} Other popular indicators include the R2 indicator~\cite{10.1145/2330163.2330230} and the Inverted Generational Distance (IGD)~\cite{Bosman2003The}. For instance, \textcolor{blue}{\mbox{MOMBI} selects offspring based on the R2 indicator~\cite{6557868}, and \mbox{MyO-DEMR} combines the concept of Pareto dominance with the IGD to select the next generation~\cite{Denysiuk2013Many}.} Moreover, the Indicator-based Evolutionary Algorithm (IBEA) can be integrated with arbitrary indicators~\cite{zitzler2004indicator}.

The algorithms in the third category, decomposition-based MaOEAs, \textcolor{blue}{decompose an MaOP into a number of single-objective optimization problems (SOPs) or simpler MOPs to be solved collaboratively. On the one hand, some decomposition-based MOEAs such as reference vector guided EA (RVEA)~\cite{2016A} and MOEA based on decomposition (MOEA/D)~\cite{zhang2007moea} decompose an MOP into a number of SOPs. On the other hand, some other decomposition-based MOEAs such as NSGA-III~\cite{Fellow2014An} and SPEA based on reference direction~\cite{2017A} decompose an MOP into several simpler MOPs by dividing the objective space into several subspaces.}

Despite many research efforts, the studies on MaOPs are still far from satisfactory. Most of the current studies focus on MaOPs with fewer than 15~objectives~\cite{10.1145/2792984,8359435}. To address this problem, we propose a new Many-Objective Evolutionary Algorithm with Determinantal Point Processes (MaOEADPPs), which performs very well on a variety of types of test instances with the number of objectives ranging from 5 to 30. Our MaOEADPPs quantitatively calculates the distribution of all solutions in the population using a Kernel Matrix. The Kernel Matrix can be split into quality and similarity components, which represent the diversity and convergence of the population, respectively. Determinantal Point Processes (DPPs) are then employed to select the solutions with higher diversity and convergence based on the decomposition of the Kernel Matrix. To further improve the performance of the Kernel Matrix, we introduce corner solutions~\cite{singh2011pareto}.

The main contributions of this paper are summarized as follows.%
\begin{enumerate}%
\item In order to adapt to different types of MaOPs and different objectives, we propose a new Kernel Matrix in which a similarity measure is defined as the cosine value of the angle between two solutions and the quality of a solution is measured by the $L_{2}$~norm of its objective vector.%
\item We propose Determinantal Point Processes Selection \mbox{(DPPs-Selection)} to select a subset of the population. We then integrate the DPPs-Selection and the Kernel Matrix to obtain the new Many-Objective Evolutionary Algorithm with Determinantal Point Processes~(MaOEADPPs).%
\end{enumerate}%

\textcolor{blue}{The rest of this paper is organized as follows.} In Section~\ref{sec:background}, we introduce the background knowledge. In Section~\ref{sec:algo}, we describe the details of our new algorithm. After that, the experimental results are given in Section~\ref{sec:experiment}. Finally, we summarize this paper and discuss the future work in Section~\ref{sec:conclusios}.%
\section{BACKGROUND}
\label{sec:background}%
\subsection{Determinantal Point Processes}%
Determinantal Point Processes (DPPs) have been used in subset selection tasks such as product recommendation~\cite{10.1145/2959100.2959178}, text summarization~\cite{kulesza2012determinantal}, web search~\cite{kulesza2011k}, and graph sampling~\cite{tremblay2017graph}.
As shown in Equation~(\ref{LL}), a point process is a probabilistic measure for one instantiation~$Y$ of a set~$\mathbf{Y}$.
We assume discrete finite point processes, i.e., $\mathbf{Y}=\{1..N\}$.
\begin{equation}
\mathcal{P}_{\mathbf{L}}(Y=\mathbf{Y})=\frac{\operatorname{det}\left(\mathbf{L}_{Y}\right)}{\operatorname{det}(\mathbf{L}+\mathbf{I})}
\label{LL}
\end{equation}%
\noindent where $\mathbf{L}$~is an $N \times N$ semi-definite Kernel Matrix indexed by~$\{1..N\}$, $\mathbf{I}$~is the $N \times N$ identity matrix, and $\rm det(\cdot)$ is the determinant.
$\mathbf{L}_{Y}$ is the Kernel Matrix $\mathbf{L}$ with indices restricted by entries indexed in the subset~$Y$. For example, if the subset is restricted to one element $Y=\{i\}$, then $\mathbf{L}_{Y}={L}_{ii}$. The element~$L_{ij}$ measures the similarity of $i$\textsuperscript{th} and $j$\textsuperscript{th} elements in the set~$\mathbf{Y}$. If we restrict the cardinality of the instantiation~$Y$ to~$k$, the Determinantal Point Processes are called \mbox{$k$-DPPs}. In this case, Equation~(\ref{hh}) holds:%
\begin{equation}
\mathcal{P}_{\mathbf{L}}^{k}(Y=\mathbf{Y})=\frac{\operatorname{det}\left(\mathbf{L}_{Y}\right)}{\sum_{\left|Y^{\prime}\right|=k} \operatorname{det}\left(\mathbf{L}_{Y^{\prime}}\right)}
\label{hh}
\end{equation}
To reduce the cost of calculating~$\mathcal{P}_{\mathbf{L}}^{k}(Y)$, which requires $O\left(N^{k}\right)$ steps, Kulesza and Taskar proposed the following method~\cite{kulesza2011k}. First, the Kernel Matrix~$\mathbf{L}$ is eigen-decomposed $\mathbf{L}=\sum_{n=1}^{N} \lambda_{n} \mathbf{v}_{n} \mathbf{v}_{n}^{T}$ with a set of eigenvectors~$\mathbf{V}$ and eigenvalues~$\lambda$. If $\mathbf{J}$~constitutes a set of $k$~selected eigenvectors and $J$~is an instantiation of~$\mathbf{J}$, the probability for $J$~can be derived as%
\begin{equation}
\operatorname{Pr}(J=\mathbf{J})=\frac{\prod_{n \in J} \lambda_{n}}{\sum_{|J|=k} \prod_{m \in J} \lambda_{m}}
\end{equation}%
In the denominator, the sum goes over all possible instantiations~$J$ of~$\mathbf{J}$.
Here $Pr$ denotes the probability, whereas $\mathcal{P}_{\mathrm{L}}(Y=\mathbf{Y})$ is a probability measure.
In~\cite{kulesza2012determinantal}, it is shown that when sampling from DPPs, the probability of an element~$i$ from~$\mathbf{Y}$, is given by%
\begin{equation}
Pr(i)=\frac{1}{\mathbf{V}} \sum_{\mathbf{v} \in \mathbf{V}}\left(\mathbf{v}^{T} e_{i}\right)^{2}
\end{equation}%
\noindent where $e_i$~is the $i$\textsuperscript{th} unit vector $e_i=(0,\dots,1,\dots,0)$, with all elements~0 except the element at index~$i$, which is~1.%
\subsection{Corner Solution}%
In the MaOPs, a majority of solutions in the population will become nondominated after several generations~\cite{li2015many}. This often causes the algorithms to lose the selection pressure. To solve this problem, we modify the definition of solution quality with the assistance of corner solutions~\cite{singh2011pareto}. \textcolor{blue}{If we get just one solution when minimizing $k$ of $M$ objectives, this solution is called a corner solution.} However, Ran and Ishibuchi point out that the corner solutions for $k>1$ are difficult to find~\cite{Ran2017A,Ishibuchi2017Performance}. \textcolor{blue}{Therefore, we will propose approximation methods to search for corner solutions in Section III}. 

%

\section{PROPOSED ALGORITHM: MaOEADPPs}%
\label{sec:algo}%
First, the general framework of MaOEADPPs is reported. Then, we will describe the \textcolor{blue}{calculation of the Kernel Matrix.} After that, the details of Environmental Selection and other important components of MaOEADPPs are explained one by one.%
\subsection{General Framework of MaOEADPPs}%
\textcolor{blue}{The pseudocode of MaOEADPPs} is given in Algorithm~\ref{alg1}. First, an initial population~$\boldsymbol{P}$ is filled with $N$~random solutions. Then, a Corner Solution Archive~$\boldsymbol{CSA}$ is extracted from~$\boldsymbol{P}$. Based on~$\boldsymbol{P}$, the ideal point~$z^*$ and the nadir point~$z^{nad}$ are initialized in \mbox{steps 3-4}, where $z_{i}^*$ and $z_{i}^{nad}$ are calculated by Equations~(\ref{*}) and~(\ref{nad}):%
\begin{equation}
\begin{aligned} z^{*} &=\left(z_{1}^{*}, z_{2}^{*}, \ldots, z_{m}^{*}\right)^{T} \\ z_{i}^{*} &=\min(\left\{ \left(f_{i}(x)\right) | x \in \boldsymbol{P}\right\}) \end{aligned}
\label{*}
\end{equation}%
\begin{equation}%
\begin{aligned} z^{nad} &=\left(z_{1}^{nad}, z_{2}^{nad}, \ldots, z_{m}^{nad}\right)^{T} \\ z_{i}^{nad} &=\max(\left\{ \left(f_{i}(x)\right) | x \in \boldsymbol{P}\right\}) \end{aligned}%
\label{nad}%
\end{equation}%
${z}_i^*$~and $z_{i}^{nad}$~are used to normalize the objective values~$f_i(x)$, as shown by Equation~(\ref{123}).
This is important because the objective functions could have \textcolor{blue}{vastly different scales.}
\begin{equation}
f_i(x)=\dfrac{f_i(x)-{z}_i^*}{z_{i}^{nad}-{z}_i^*}.
\label{123}%
\end{equation}%
Next, \mbox{steps 5-12} are iterated until the number of function evaluations exceeds a fixed limit (100'000 in our experiments).
At each iteration, $2N$~solutions are selected as the mating pool~$\boldsymbol{P}'$.
Simulated binary crossover (SBX)~\cite{agrawal1995simulated} and polynomial mutation~\cite{deb1996combined} are employed to generate the candidate population~$\boldsymbol{C}$.
Then the ideal point~$\boldsymbol{z}^*$ is updated by Equation~(\ref{*}). After Environmental Selection, $z^{nad}$~will be updated by Equation~(\ref{nad}).%
\begin{algorithm}
  \caption{Framework of the proposed MaOEADPPs}
  \label{alg1}
  \begin{algorithmic}[1]
    \REQUIRE $ N $ (population size)
    \STATE $\boldsymbol{P} \leftarrow \ $Initialize Population($ N $)
    \STATE $\boldsymbol{CSA}$$ \leftarrow \ $$\boldsymbol{P}$
    \STATE $\boldsymbol{z}^* \leftarrow \ $Initialize Ideal Point($\boldsymbol{P}$)
    \STATE $\boldsymbol{z}^{nad} \leftarrow \ $Initialize Nadir Point($\boldsymbol{P}$)
    \WHILE{the number of function evaluations does not exceed 100'000}
    \STATE  $\boldsymbol{P}'\leftarrow $Filling the Mating Pool($\boldsymbol{P},\boldsymbol{CSA},N,\boldsymbol{z}^*,\boldsymbol{z}^{nad} $)
    \STATE  $\boldsymbol{C} \leftarrow$Variation($ \boldsymbol{P}' $)
    \STATE  $\boldsymbol{z}^* \leftarrow \ $Update Ideal Point($ \boldsymbol{z}^*,\boldsymbol{C}$)
    \STATE  $ \boldsymbol{CSA}$$ \leftarrow $UpdateCSA($\boldsymbol{CSA}, \boldsymbol{C},N$)
    \STATE  $\boldsymbol{P}$$ \leftarrow $Environmental Selection($ \boldsymbol{P} , \boldsymbol{C}, N, \boldsymbol{CSA}, \boldsymbol{z}^*,\boldsymbol{z}^{nad} $)
    \STATE  $\boldsymbol{z}^{nad} \leftarrow \ $Update Nadir Point($ \boldsymbol{P} , \boldsymbol{CSA}$)
    \ENDWHILE
    \ENSURE$\boldsymbol{P}$
  \end{algorithmic}
\end{algorithm}%
\subsection{Filling the Mating Pool}%
To produce high-quality offspring, we generate the mating pool~$\boldsymbol{P}'$ by selecting elements from the union of~$\boldsymbol{P}$ and~$\boldsymbol{CSA}$.
We therefore define the convergence~$con(x)$ of a solution~$x \in \boldsymbol{P}$ according to Equation~(\ref{222}):%
\begin{equation}
con(x)={\frac{1}{\sum_{i=1}^{m}(f_i(x))^2}}
\label{222}
\end{equation}%
Then, as illustrated in Algorithm~\ref{alg3}, we randomly choose a solution~$x$ from the union $\boldsymbol{P} \cup \boldsymbol{CSA}$.
If the convergence~$con(y)$ of solution~$y=\underset{y \in \boldsymbol{P}}{\arg \min }\{\cos (x, y)\}$ is greater than~$con(x)$ and a randomly generated number is less than a threshold~$ \delta(x,y) $, $y$~is added to~$\boldsymbol{P}'$.
Otherwise, $x$~is added to~$\boldsymbol{P}'$.
The threshold~$ \delta(x,y) $ is computed according to Equation~(\ref{333}).%
\begin{equation}
\begin{split}
 &\delta(x,y) =\frac{\cos(x,y)- minCos}{maxCos-minCos}\\
 &maxCos =  \max \limits_{p,q \in \boldsymbol{P} \cup \boldsymbol{CSA},  p \neq q}\{\cos(p,q)\}\\
 &minCos =  \min \limits_{p,q \in \boldsymbol{P} \cup \boldsymbol{CSA},  p \neq q}\{\cos(p,q)\}
\end{split}
\label{333}
\end{equation}%
Each time offspring needs to be generated, two solutions are randomly selected from mating pool~$\boldsymbol{P}'$ as the parents.%
\begin{algorithm}
  \caption{Filling the Mating Pool}
  \label{alg3}
  \begin{algorithmic}[1]
    \REQUIRE $\boldsymbol{P}$ (population), $ \boldsymbol{CSA} $ (Corner Solution Archive), $ N $ (population size), $z^*$ (ideal point), $z^{nad}$ (nadir point)

    \STATE Normalization( $\boldsymbol{P} \cup \boldsymbol{CSA} $, $z^*$, $z^{nad}$)

    \STATE  $\boldsymbol{P}' \leftarrow \emptyset$
    \FOR{$i = 1 \to 2N$}
    \STATE Randomly select $x$ from  $\boldsymbol{P} \cup \boldsymbol{CSA}$
    \STATE Find solution $y=\underset{y \in \boldsymbol{P}}{\arg \min }\{\cos (x, y)\}$
    \IF{$ (rand <  \delta  \wedge ({con}(y) > {con}(x)) $}
    \STATE $ \boldsymbol{P}' $$\leftarrow $$ \boldsymbol{P}' \bigcup \{y\} $
    \ELSE
    \STATE $ \boldsymbol{P}' $$\leftarrow $$ \boldsymbol{P}' \bigcup \{x\} $
    \ENDIF

    \ENDFOR
    \ENSURE $\boldsymbol{P}' $

  \end{algorithmic}
\end{algorithm}%
\subsection{Environmental Selection}%
\begin{algorithm}
  \caption{Environmental Selection}
  \label{alg4}
  \begin{algorithmic}[1]
    \REQUIRE $\boldsymbol{P}$ (population), $\boldsymbol{C}$ (candidate population), $N$ (population size), $\boldsymbol{CSA}$ (Corner Solution Archive), $z^*$ (ideal point), $z^{nad}$ (nadir point)
    \STATE $P \leftarrow \ $Nondominated($\boldsymbol{P} \cup \boldsymbol{C}$)
    \IF{$|\boldsymbol{P}| > N$}
    \STATE Normalization($\boldsymbol{CSA}$, $z^*$, $z^{nad}$)
    \STATE Normalization($\boldsymbol{P}$, $\boldsymbol{z}^*$, $z^{nad}$)
    \STATE $\boldsymbol{L}$$\leftarrow$Calculation of the Kernel Matrix($\boldsymbol{P}$, $\boldsymbol{CSA}$)
    \STATE $ \boldsymbol{A} $$ \leftarrow $DPPs-Selection($ \boldsymbol{L} $, $ N $)
    \STATE $ \boldsymbol{P} \leftarrow  \boldsymbol{P}(\boldsymbol{A}) $
    \ENDIF
    \ENSURE $\boldsymbol{P}$
  \end{algorithmic}
\label{ES}
\end{algorithm}%
The framework of Environmental Selection is given in Algorithm~\ref{ES}. First, the nondominated solutions are selected from the union of~$\boldsymbol{P}$ and~$\boldsymbol{C}$. If the number of nondominated solutions is greater than~$N$, we calculate the Kernel Matrix~$\boldsymbol{L}$ and update~$\boldsymbol{P}$ with DPPs-Selection.

The Kernel Matrix~$\boldsymbol{L}$ is defined to represent both the diversity and the convergence of the population. We use Equation~(\ref{888}) to calculate its elements~${L}_{xy}$:%
\begin{equation}
{L}_{xy}=q(x)S(x,y)q(y)
\label{888}
\end{equation}%
\noindent where $x, y\in \boldsymbol{P} $, $q(x)$~is the quality of solution~$x$, and $S(x,y)$~is the similarity between elements~$x $ and~$y$.
To measure the similarity between~$x$ and~$y$, $S(x,y)$~is defined by Equation~(\ref{444}).%
\begin{equation}
S(x,y)=\exp{(-\cos(x,y))}%
\label{444}%
\end{equation}%
\noindent where~$ \cos (x,y) $ is the cosine of the angle between solutions~$x$ and~$y$.
The quality~$q(x)$ of solution~$x$ is calculated based on its convergence, as shown by Equations~(\ref{777}) and~(\ref{666}).%
\begin{equation}
q(x)=
\begin{cases}
{con}_1(x)& x\in outside \: space \\
2*\max \limits_{p \in \boldsymbol{P}}({con}_1(p))& x \in inside \:space
\end{cases}
\label{777}
\end{equation}%
\begin{equation}%
{con}_1(x)=\dfrac{con(x)}{\max \limits_{p \in \boldsymbol{P}}(con(p))}
\label{666}
\end{equation}%
\noindent where~${con}_1$ is the normalized convergence. $outside\ space$ and $inside\ space$ describe the different areas of the objective space. If $\sqrt{\sum_{i=1}^{M} f_{i}(x)^{2}}\leq\operatorname{t}$, the solution~$x$ belongs to the $inside\ space$ and otherwise to the $outside\ space$. The threshold~$t$ is set to \mbox{$t=\max{\left\{ \left. \sqrt{\sum_{i=1}^{M} f_{i}(x)^{2}}\right| x \in \boldsymbol{CSA}\right\}}$}.

The Corner Solution Archive~$\boldsymbol{CSA}$ is used to distinguish the $outside\ space$ and $inside\ space$. Since it is difficult to find the corner solutions for $k>1$~\cite{Ran2017A,Ishibuchi2017Performance}, we use an approximation method to generate the ~$\boldsymbol{CSA}$. According to the value of $k$, we discuss the corner solutions in two cases.

$\boldsymbol{k=1}$: To\ \  find\ \  the\ \  corner\ \  solutions\ \  of\ \  objective

\noindent$i=1,2,...,M$, we sort the solutions in ascending order of the objective value~$f_i$, so we have $M$~sorted lists and we add the first $\lceil \frac{N}{3M}\rceil$ solutions of each list into the $\boldsymbol{CSA}$.

$\boldsymbol{1<k< M}$: We only consider $k=M - 1$ and employ an approximation method to get the $\boldsymbol{CSA}$. For any objective $i=1,2,...,M$, we sort the solutions in ascending order of $ \sqrt{\sum_{j=1,j \neq i}^{M}(f_j(x))^2} $ and obtain $M$~sorted lists. The first $\lceil \frac{2N}{3M}\rceil$~solutions of each list are selected into the $\boldsymbol{CSA}$.

From \quad the \quad above \quad two \quad cases, \quad we \quad get $|\boldsymbol{CSA}|=\lceil\frac{N}{3M}\rceil \times M+\lceil\frac{2N}{3M}\rceil \times M \approx N$. That is each objective contributes $\lceil \frac{N}{M}\rceil$ solutions to the $\boldsymbol{CSA}$.

Algorithm~\ref{alg5} shows the details of calculating the Kernel Matrix. After calculating the cosine of the angle between every two solutions in the population, the quality~$q(x)$ of each solution~$x$ is computed. The row vector~$\boldsymbol{q}$ then contains the qualities of all solutions. After that, a quality matrix~$\boldsymbol{Q}$ is generated as the product of~$\boldsymbol{q}^T$ and~$\boldsymbol{q}$. Finally, $\boldsymbol{Q}$ is multiplied with~$\boldsymbol{L}$ in element-wised manner to update and output~$\boldsymbol{L}$. In \mbox{step 10}, `$\otimes$' indicates the element-wised multiplication of two matrices with the same size.%
\begin{algorithm}
  \caption{Calculation of the Kernel Matrix}
  \label{alg5}
  \begin{algorithmic}[1]
    \REQUIRE  $ \boldsymbol{P}$ (population), $\boldsymbol{CSA}$ (Corner Solution Archive)
    \FOR{solution $x \in  \boldsymbol{P}$}
    \FOR{solution $y \in  \boldsymbol{P}$}
    \STATE $\boldsymbol{L}_{xy}=cos(x,y)$
    \ENDFOR
    \ENDFOR
    \FOR{solution $x \in  \boldsymbol{P}$}
    \STATE calculate $q(x)$ by Equations (\ref{777}) and (\ref{666})
    \ENDFOR
    \STATE $\boldsymbol{Q}=\boldsymbol{q}^T \cdot \boldsymbol{q}$
    \STATE $\boldsymbol{L}=\boldsymbol{L}\otimes \boldsymbol{Q}$
    \ENSURE $\boldsymbol{L}$
  \end{algorithmic}
\end{algorithm}

In Algorithm~\ref{DPPS}, we specify the details of the Determinantal Point Processes Selection (DPPs-Selection). First, the $N\times N$ Kernel Matrix $\boldsymbol{L}$ is eigen-decomposed to obtain the eigenvector set $\boldsymbol{V}=\{\boldsymbol{v}_{r}\}^{N}_{r=1}  $ and the eigenvalue set~$\boldsymbol{\lambda}=\{\lambda_{r}\}^{N}_{r=1}  $.
Then, $ \{\boldsymbol{v}_{r}\}^{N}_{r=1} $~is sorted in decreasing order of the eigenvalues and is truncated by keeping the $k$~largest eigenvectors only.
After that, on each iteration of the while-loop, the index of the element that has the maximum $\sum_{\boldsymbol{v}\in \boldsymbol{V}}(\boldsymbol{v}^\top \boldsymbol{e}_i) ^2 $ is added into the index set~$\boldsymbol{S}$, where the dimension of~$\boldsymbol{e}_i$ is~$N$ and $\boldsymbol{e}_i=(0,0,\dots,1,\dots,0,0)$ is the $i$\textsuperscript{th} standard unit vector in $\mathbb{R}^N$. Then $\boldsymbol{V}$~is replaced by an orthonormal basis for the subspace of $\boldsymbol{V}$~orthogonal to~$\boldsymbol{e}_{i}$. Finally, the index set~$\boldsymbol{S}$ of the selected elements is returned. In summary, DPPs-Selection has two phases: It first selects $k$~eigenvectors of~$\boldsymbol{L}$ that have the maximum corresponding~$\lambda$. Then, it selects the indices of elements that have the maximum $\sum_{\boldsymbol{v}\in \boldsymbol{V}}(\boldsymbol{v^\top} \boldsymbol{e}_i) ^2 $ one by one.%
\begin{algorithm}
  \caption{DPPs-Selection}
  \label{alg2}
  \begin{algorithmic}[1]
    \REQUIRE  $ \boldsymbol{L}$(Kernel Matrix), $k$ (size of subset)

    \STATE $(\{\boldsymbol{v}_{r}\}^{N}_{r=1},\{\lambda_{r}\}^{N}_{r=1}) \leftarrow $Eigen-decomposition($ \boldsymbol{L} $)

    \STATE  $\boldsymbol{V} \leftarrow $Select $  k $ eigenvectors from $\{\boldsymbol{v}_{r}\}^{N}_{r=1} $ according to the largest $  k $ eigenvalues in $ \{\lambda_{r}\}^{N}_{r=1} $
    \STATE  $\boldsymbol{S} \leftarrow \phi $
    \WHILE{$| \boldsymbol{V}|>0$}
    \STATE Select $i$ from $N$ that has maximum $ \sum_{\boldsymbol{v}\in \boldsymbol{V}}^{}(\boldsymbol{v^\top e}_i) ^2  $
    \STATE $\boldsymbol{S}\leftarrow \boldsymbol{S}\cup i$
    \STATE $\boldsymbol{V}  \leftarrow  \boldsymbol{V}_\perp $, an orthonormal basis for the subspace of $ \boldsymbol{V} $ orthogonal to $\boldsymbol{e}_{i}$
    \ENDWHILE
    \ENSURE  $ \boldsymbol{S}$
  \end{algorithmic}
  \label{DPPS}
\end{algorithm}%
\section{EXPERIMENTAL STUDY}%
\label{sec:experiment}%
To evaluate the performance of~MaOEADPPs, we compare it with the state-of-the-art algorithms \mbox{NSGA-III~\cite{deb2014evolutionary}}, RSEA~\cite{he2017radial}, \mbox{$\theta $-DEA~\cite{yuan2016new}}, \mbox{AR-MOEA~\cite{tian2017indicator}}, and RVEA~\cite{cheng2016reference} on a set of benchmark problems with 5, 10, 13, 15, 20, 25, and 30~objective functions. After that, we investigate the effectiveness of our new Kernel Matrix and DPPs-Selection.
All the experiments are conducted using PlatEMO~\cite{tian2017platemo}, which is an evolutionary multi-objective optimization platform.%
\subsection{Test Problems and Performance Indicators}%
To evaluate our methods under a wide variety of different problems types, we choose the benchmarks \mbox{DTLZ1-DTLZ6~\cite{deb2005scalable}}, IDTLZ1 and IDTLZ2~\cite{jain2014evolutionary}, \mbox{WFG1-WFG9~\cite{huband2006review}}, and \mbox{MAF1-MAF7~\cite{Ran2017A}}.

WFG1 is a separable and unimodal problem.
WFG2 has a scaled-disconnected PF and WFG3 has a linear PF.
WFG4 introduces multi-modality to test the ability of algorithms to jump out of local optima.
WFG5 is a highly deceptive problem.
The nonseparable reduction of WFG6 is difficult.
WFG7 is a separable and unimodal problem.
WFG8 and WFG9 introduce significant bias to evaluate the MaOEA performance.

DTLZ1 is a simple test problem with a triangular~PF. DTLZ2 to DTLZ4 have concave~PFs. DTLZ5 tests the ability of an MaOEA to converge to a curve. DTLZ6 is a modified version of DTLZ5. IDTLZ1 and IDTLZ2 are obtained by inverting the~PFs of DTLZ1 and DTLZ2, respectively.

MaF1 is a modified inverted instance of DTLZ1. MaF2 is modified from DTLZ2 to increase the difficulty of convergence. MaF3 is a convex version of DTLZ3. MaF4 is used to assess a MaOEA performance on problems with badly scaled PFs. MaF5 has a highly biased solution distribution. MaF6 is used to assess whether MaOEAs are capable of dealing with degenerate PFs. MaF7 is used to assess MaOEAs on problems with disconnected PFs.

\textcolor{blue}{Following the settings in the original papers~\cite{deb2005scalable,jain2014evolutionary,huband2006review,Ran2017A}, for DTLZ2-DTLZ4, MaF1-MaF7, IDTLZ2 and WFG1-WFG9, the numbers of decision variables~$D$ are set to \mbox{$M-1+10$}, where $M$~denotes the number of objectives. For IDTLZ1 and DTLZ1, the numbers of decision variables~$D$ are set to \mbox{$M-1+5$}.}
As performance metrics, we adopt the widely used IGD~\cite{zhou2006combining } and HV~\cite{huband2006review} indicators.%
\subsection{Experimental Settings}%
Population size: The population size in our experiments is the same as the number of reference points.
The reference points are generated by Das and Dennis's approach~\cite{das1998normal} with two layers.
This method is adapted in \mbox{NSGA-III}, \mbox{$ \theta $-DEA}, and RVEA.
For fair comparisons, the population size of all algorithms is set according to Table~\ref{No Label}. According to the above method, there is no appropriate parameter to ensure that the population size for 13~objectives is less than for 15~objectives and greater than for 10~objectives. We thus set the population size of 13~objectives to be the same as 15~objectives.

Genetic operators: we apply simulated binary crossover (SBX)~\cite{agrawal1995simulated} and polynomial mutation~\cite{deb1996combined} to generate the offspring.
The crossover probability~$p_{c}$ and the mutation probability~$p_{m}$ are set to~1.0 and~$1/D$, respectively, where~$D$ denotes the number of decision variables.
Both distribution indices of crossover and mutation are set to~20.%
\begin{table}[tb]
  \renewcommand{\arraystretch}{1.2}
  \centering
  \caption{POPULATION SIZE SETTINGS}
    \begin{tabular}{ccc}
      \toprule
      Number of objectives&Parameter&Population size\\
      (M)&($p_1$,$p_2$)&(N)\\
      \midrule
      \multirow{1}{*}{5}&{(5,0)} &126\\
      \multirow{1}{*}{10}&{(3,1)}&230\\
      \multirow{1}{*}{15}&{(2,2)}&240\\
      \bottomrule
    \end{tabular}
    \label{No Label}
  \end{table}

Termination condition and the number of runs: 
The tested algorithms will be terminated when the number of function evaluations exceeds~100'000. All algorithms are executed for 30~independent runs on each test instance.

Parameter settings for algorithms:
The two parameters $\alpha$ and $f_r$ of RVEA are set to~2 and~0.1, respectively, which is the same as in~\cite{cheng2016reference}.%
\subsection{Experimental Results and Analysis}%
We now analyze the results of the aforementioned state-of-the-art algorithms on our benchmark instances. The mean values and the standard deviations of the IGD and HV metrics on DTLZ1-DTLZ6, IDTLZ1-IDTLZ2 and WFG1-WFG9 with~5, 10, 13, and~15 objectives are listed in Tables~\ref{19IGD} and~\ref{19HV}, respectively.
In each table, the best mean value for each instance is highlighted. The symbols `+', `-', and `$\approx$' indicate whether the results are significantly better than, significantly worse than, or not significantly different from those obtained by MaOEADPPs, respectively. We apply the Wilcoxon rank sum test with a significance level of~0.05.%
\begin{table*}[tbp]
  \renewcommand{\arraystretch}{1.2}
  \centering
  \caption{IGD VALUES OBTAINED BY MAOEADPPS, \mbox{AR-MOEA~\cite{tian2017indicator}}, \mbox{NSGA-III~\cite{deb2014evolutionary}}, \mbox{RSEA~\cite{he2017radial}}, \mbox{$\theta $-DEA~\cite{yuan2016new}} AND RVEA~\cite{cheng2016reference} ON \mbox{DTLZ1-DTLZ6}, \mbox{IDTLZ1-IDTLZ2} AND \mbox{WFG1-WFG9} WITH 5, 10, 13, AND 15 OBJECTIVES. THE BEST RESULTS FOR EACH INSTANCE ARE HIGHLIGHTED IN GRAY.}
  \resizebox{0.96\textwidth}{!}{
    \begin{tabular}{cccccccc}
      \toprule
      Problem&$M$&MaOEADPPs&AR-MOEA&\mbox{NSGA-III}&RSEA&$\theta $-DEA&RVEA\\
      \midrule
      \multirow{1}{*}{DTLZ1}&5&\hl{6.3306e-2 (2.01e-3)}&6.3325e-2 (8.43e-5) $-$&6.3377e-2 (6.27e-5) $-$&7.6739e-2 (3.13e-3) $-$&6.3356e-2 (7.85e-5) $-$&6.3322e-2 (4.56e-5) $-$\\
      \multirow{1}{*}{DTLZ1}&10&\hl{1.1196e-1 (1.05e-3)}&1.3914e-1 (3.67e-3) $-$&1.7758e-1 (8.00e-2) $-$&1.3925e-1 (7.37e-3) $-$&1.3572e-1 (2.74e-2) $-$&1.3167e-1 (2.59e-3) $-$\\
      \multirow{1}{*}{DTLZ1}&13&1.3318e-1 (2.66e-3)&\hl{1.3186e-1 (3.36e-3) $+$}&1.8794e-1 (5.80e-2) $-$&1.8854e-1 (2.71e-2) $-$&1.3913e-1 (3.37e-2) $-$&1.3376e-1 (1.24e-2) $-$\\
      \multirow{1}{*}{DTLZ1}&15&1.3042e-1 (2.51e-3)&\hl{1.2670e-1 (2.49e-3) $+$}&2.2759e-1 (1.11e-1) $-$&1.8914e-1 (1.88e-2) $-$&1.3793e-1 (3.50e-2) $-$&1.3525e-1 (1.25e-2) $\approx$\\
      \hline
      \multirow{1}{*}{DTLZ2}&5&\hl{1.9245e-1 (9.58e-4)}&1.9485e-1 (3.23e-5) $-$&1.9489e-1 (6.18e-6) $-$&2.4892e-1 (1.97e-2) $-$&1.9488e-1 (4.99e-6) $-$&1.9489e-1 (6.09e-6) $-$\\
      \multirow{1}{*}{DTLZ2}&10&\hl{4.1972e-1 (3.55e-3)}&4.4256e-1 (2.02e-3) $-$&5.0385e-1 (7.71e-2) $-$&5.2741e-1 (2.66e-2) $-$&4.5439e-1 (7.45e-4) $-$&4.5336e-1 (4.27e-4) $-$\\
      \multirow{1}{*}{DTLZ2}&13&\hl{4.8761e-1 (1.52e-3)}&5.1536e-1 (1.26e-3) $-$&6.5217e-1 (6.64e-2) $-$&6.1696e-1 (3.21e-2) $-$&5.1595e-1 (5.08e-4) $-$&5.1725e-1 (4.39e-4) $-$\\
      \multirow{1}{*}{DTLZ2}&15&\hl{5.2592e-1 (1.75e-3)}&5.2956e-1 (1.44e-3) $-$&6.6720e-1 (7.64e-2) $-$&6.5998e-1 (2.31e-2) $-$&5.2681e-1 (7.00e-4) $\approx$&5.2688e-1 (7.94e-4) $\approx$\\
      \hline
      \multirow{1}{*}{DTLZ3}&5&2.0020e-1 (3.76e-3)&1.9587e-1 (1.02e-3) $+$&1.9646e-1 (2.05e-3) $+$&2.5767e-1 (4.25e-2) $-$&\hl{1.9556e-1 (6.73e-4) $+$}&2.2695e-1 (1.63e-1) $-$\\
      \multirow{1}{*}{DTLZ3}&10&\hl{4.1885e-1 (5.64e-3)}&4.9847e-1 (1.74e-1) $-$&5.5397e+0 (6.08e+0) $-$&7.7705e-1 (2.15e-1) $-$&5.6241e-1 (2.36e-1) $-$&4.6015e-1 (8.78e-3) $-$\\
      \multirow{1}{*}{DTLZ3}&13&\hl{4.9278e-1 (1.23e-2)}&5.3091e-1 (7.92e-3) $-$&2.8251e+1 (1.35e+1) $-$&1.1572e+0 (5.99e-1) $-$&5.8925e-1 (8.13e-2) $-$&5.2137e-1 (4.99e-3) $-$\\
      \multirow{1}{*}{DTLZ3}&15&\hl{5.2836e-1 (1.01e-2)}&5.5077e-1 (8.21e-3) $-$&5.7390e+1 (3.45e+1) $-$&2.4499e+0 (1.59e+0) $-$&6.6445e-1 (3.00e-1) $-$&6.0578e-1 (2.22e-1) $-$\\
      \hline
      \multirow{1}{*}{DTLZ4}&5&\hl{1.9296e-1 (8.72e-4)}&1.9489e-1 (4.24e-5) $-$&2.2981e-1 (8.54e-2) $-$&2.5669e-1 (2.52e-2) $-$&1.9488e-1 (2.11e-5) $-$&2.0326e-1 (4.35e-2) $-$\\
      \multirow{1}{*}{DTLZ4}&10&\hl{4.2861e-1 (3.04e-3)}&4.4576e-1 (2.25e-3) $-$&4.8641e-1 (6.16e-2) $-$&5.3888e-1 (1.11e-2) $-$&4.5616e-1 (5.96e-4) $-$&4.5612e-1 (4.20e-4) $-$\\
      \multirow{1}{*}{DTLZ4}&13&\hl{4.9186e-1 (8.20e-4)}&5.1932e-1 (1.04e-3) $-$&5.9984e-1 (7.22e-2) $-$&6.0774e-1 (1.51e-2) $-$&5.1459e-1 (8.28e-4) $-$&5.2260e-1 (1.55e-2) $-$\\
      \multirow{1}{*}{DTLZ4}&15&\hl{5.2851e-1 (8.09e-4)}&5.4819e-1 (1.61e-3) $-$&6.1083e-1 (7.51e-2) $-$&6.6149e-1 (1.44e-2) $-$&5.2943e-1 (1.59e-3) $\approx$&5.2963e-1 (1.53e-3) $-$\\
      \hline
      \multirow{1}{*}{DTLZ5}&5&5.3392e-2 (1.29e-3)&6.8178e-2 (1.16e-2) $-$&3.0807e-1 (2.64e-1) $-$&\hl{3.5076e-2 (6.23e-3) $+$}&2.2564e-1 (8.54e-2) $-$&1.8215e-1 (2.03e-2) $-$\\
      \multirow{1}{*}{DTLZ5}&10&\hl{9.2301e-2 (7.45e-3)}&1.2055e-1 (3.01e-2) $-$&4.0553e-1 (1.34e-1) $-$&1.1613e-1 (4.65e-2) $\approx$&1.7127e-1 (2.82e-2) $-$&2.3915e-1 (1.90e-2) $-$\\
      \multirow{1}{*}{DTLZ5}&13&8.8231e-2 (1.08e-2)&\hl{7.7819e-2 (9.77e-3) $+$}&9.2041e-1 (2.44e-1) $-$&9.7573e-2 (4.83e-2) $\approx$&1.9347e-1 (3.24e-2) $-$&6.2786e-1 (1.81e-1) $-$\\
      \multirow{1}{*}{DTLZ5}&15&1.3319e-1 (1.81e-2)&1.0924e-1 (3.22e-2) $+$&8.8813e-1 (2.62e-1) $-$&\hl{8.5798e-2 (2.54e-2) $+$}&1.8303e-1 (3.35e-2) $-$&6.3540e-1 (2.00e-1) $-$\\
      \hline
      \multirow{1}{*}{DTLZ6}&5&\hl{5.7741e-2 (3.06e-3)}&8.3514e-2 (1.93e-2) $-$&4.2988e-1 (3.84e-1) $-$&8.2147e-2 (3.12e-2) $-$&2.8968e-1 (1.06e-1) $-$&1.3080e-1 (2.32e-2) $-$\\
      \multirow{1}{*}{DTLZ6}&10&\hl{7.5953e-2 (6.23e-3)}&1.0986e-1 (3.35e-2) $-$&4.2037e+0 (1.33e+0) $-$&1.9300e-1 (7.82e-2) $-$&2.8761e-1 (5.93e-2) $-$&2.2295e-1 (6.91e-2) $-$\\
      \multirow{1}{*}{DTLZ6}&13&9.4493e-2 (9.89e-3)&\hl{8.3196e-2 (1.52e-2) $\approx$}&5.2170e+0 (9.12e-1) $-$&2.2475e-1 (1.17e-1) $-$&2.8129e-1 (7.97e-2) $-$&2.9068e-1 (1.05e-1) $-$\\
      \multirow{1}{*}{DTLZ6}&15&1.1107e-1 (1.07e-2)&\hl{1.0497e-1 (6.45e-2) $+$}&6.3554e+0 (1.05e+0) $-$&2.4628e-1 (9.49e-2) $-$&2.5667e-1 (7.33e-2) $-$&3.0595e-1 (1.28e-1) $-$\\
      \hline
      \multirow{1}{*}{WFG1}&5&4.2909e-1 (5.21e-3)&4.5288e-1 (7.92e-3) $-$&4.4764e-1 (1.19e-2) $-$&4.7135e-1 (1.53e-2) $-$&\hl{4.1655e-1 (8.63e-3) $+$}&4.4311e-1 (9.34e-3) $-$\\
      \multirow{1}{*}{WFG1}&10&\hl{1.0037e+0 (2.04e-2)}&1.3512e+0 (8.43e-2) $-$&1.3132e+0 (8.82e-2) $-$&1.0870e+0 (3.43e-2) $-$&1.0216e+0 (2.38e-2) $-$&1.2741e+0 (5.01e-2) $-$\\
      \multirow{1}{*}{WFG1}&13&\hl{1.4412e+0 (3.84e-2)}&2.1082e+0 (6.78e-2) $-$&1.8972e+0 (1.02e-1) $-$&1.6079e+0 (3.47e-2) $-$&1.5595e+0 (3.85e-2) $-$&1.7786e+0 (5.98e-2) $-$\\
      \multirow{1}{*}{WFG1}&15&\hl{1.4530e+0 (4.63e-2)}&1.9721e+0 (6.32e-2) $-$&1.9674e+0 (6.76e-2) $-$&1.5967e+0 (2.59e-2) $-$&1.5368e+0 (2.52e-2) $-$&1.7273e+0 (7.46e-2) $-$\\
      \hline
      \multirow{1}{*}{WFG2}&5&4.6383e-1 (7.54e-3)&4.7438e-1 (3.05e-3) $-$&4.7243e-1 (1.68e-3) $-$&4.9592e-1 (1.44e-2) $-$&4.5444e-1 (3.46e-3) $+$&\hl{4.4525e-1 (1.01e-2) $+$}\\
      \multirow{1}{*}{WFG2}&10&1.2048e+0 (6.74e-2)&1.1139e+0 (4.85e-2) $+$&1.4527e+0 (1.92e-1) $-$&\hl{1.0867e+0 (3.97e-2) $+$}&1.5145e+0 (2.39e-1) $-$&1.1270e+0 (2.62e-2) $+$\\
      \multirow{1}{*}{WFG2}&13&1.7090e+0 (4.39e-2)&\hl{1.6004e+0 (3.30e-2) $+$}&2.0962e+0 (1.06e-1) $-$&1.6762e+0 (8.96e-2) $\approx$&3.9204e+0 (9.92e-1) $-$&1.8705e+0 (1.11e-1) $-$\\
      \multirow{1}{*}{WFG2}&15&1.5696e+0 (4.51e-2)&\hl{1.5117e+0 (4.07e-2) $+$}&2.0266e+0 (1.03e-1) $-$&1.9711e+0 (2.15e-1) $-$&4.2596e+0 (1.59e+0) $-$&1.6877e+0 (1.01e-1) $-$\\
      \hline
      \multirow{1}{*}{WFG3}&5&4.3028e-1 (2.57e-2)&5.5961e-1 (3.51e-2) $-$&5.6247e-1 (5.52e-2) $-$&\hl{1.5982e-1 (4.00e-2) $+$}&4.8648e-1 (6.30e-2) $-$&5.3706e-1 (2.50e-2) $-$\\
      \multirow{1}{*}{WFG3}&10&1.4240e+0 (6.56e-2)&2.6322e+0 (1.31e-1) $-$&1.4317e+0 (7.37e-1) $\approx$&\hl{3.1341e-1 (5.56e-2) $+$}&1.1024e+0 (1.35e-1) $+$&3.7593e+0 (9.31e-1) $-$\\
      \multirow{1}{*}{WFG3}&13&2.6913e+0 (2.48e-1)&4.0574e+0 (2.70e-1) $-$&2.9959e+0 (1.04e+0) $\approx$&2.1943e+0 (1.38e+0) $\approx$&\hl{2.0254e+0 (3.81e-1) $+$}&5.9965e+0 (1.68e+0) $-$\\
      \multirow{1}{*}{WFG3}&15&3.1423e+0 (2.22e-1)&4.7374e+0 (2.67e-1) $-$&4.7042e+0 (2.86e+0) $-$&2.8008e+0 (1.68e+0) $\approx$&\hl{2.2042e+0 (2.29e-1) $+$}&7.2636e+0 (1.40e+0) $-$\\
      \hline
      \multirow{1}{*}{WFG4}&5&\hl{1.1159e+0 (8.92e-3)}&1.1762e+0 (7.50e-4) $-$&1.1844e+0 (4.02e-2) $-$&1.3049e+0 (3.56e-2) $-$&1.1770e+0 (8.12e-4) $-$&1.1771e+0 (1.08e-3) $-$\\
      \multirow{1}{*}{WFG4}&10&\hl{4.2115e+0 (4.16e-2)}&4.7923e+0 (1.82e-2) $-$&4.7963e+0 (7.18e-2) $-$&4.7826e+0 (8.57e-2) $-$&4.7729e+0 (1.11e-2) $-$&4.6176e+0 (6.09e-2) $-$\\
      \multirow{1}{*}{WFG4}&13&\hl{6.0411e+0 (3.45e-2)}&7.6982e+0 (2.39e-1) $-$&7.5013e+0 (3.21e-1) $-$&7.2177e+0 (1.25e-1) $-$&7.4910e+0 (2.85e-2) $-$&7.4632e+0 (2.15e-1) $-$\\
      \multirow{1}{*}{WFG4}&15&\hl{7.4994e+0 (9.74e-2)}&9.1459e+0 (1.28e-1) $-$&8.7909e+0 (4.67e-1) $-$&9.2549e+0 (3.36e-1) $-$&8.0876e+0 (3.42e-2) $-$&8.9541e+0 (1.52e-1) $-$\\
      \hline
      \multirow{1}{*}{WFG5}&5&\hl{1.1002e+0 (6.85e-3)}&1.1643e+0 (3.05e-4) $-$&1.1647e+0 (3.26e-4) $-$&1.3163e+0 (4.52e-2) $-$&1.1645e+0 (4.44e-4) $-$&1.1655e+0 (5.87e-4) $-$\\
      \multirow{1}{*}{WFG5}&10&\hl{4.5244e+0 (3.65e-2)}&4.7673e+0 (1.66e-2) $-$&4.7239e+0 (7.25e-3) $-$&4.8274e+0 (9.09e-2) $-$&4.7244e+0 (8.72e-3) $-$&4.6519e+0 (5.33e-2) $-$\\
      \multirow{1}{*}{WFG5}&13&\hl{6.0729e+0 (2.83e-2)}&7.2306e+0 (1.68e-1) $-$&7.5149e+0 (2.29e-1) $-$&7.2262e+0 (2.00e-1) $-$&7.3677e+0 (5.26e-2) $-$&7.2965e+0 (1.93e-1) $-$\\
      \multirow{1}{*}{WFG5}&15&\hl{7.1786e+0 (7.40e-2)}&8.0269e+0 (2.18e-1) $-$&8.1333e+0 (3.13e-1) $-$&8.8997e+0 (1.49e-1) $-$&7.8561e+0 (8.51e-2) $-$&8.7826e+0 (1.56e-1) $-$\\
      \hline
      \multirow{1}{*}{WFG6}&5&\hl{1.1186e+0 (9.54e-3)}&1.1629e+0 (1.63e-3) $-$&1.1631e+0 (2.04e-3) $-$&1.3048e+0 (3.49e-2) $-$&1.1630e+0 (2.13e-3) $-$&1.1644e+0 (2.18e-3) $-$\\
      \multirow{1}{*}{WFG6}&10&\hl{4.4049e+0 (5.41e-2)}&4.8057e+0 (1.07e-2) $-$&5.4555e+0 (1.11e+0) $-$&4.9602e+0 (9.47e-2) $-$&4.7680e+0 (9.39e-3) $-$&4.4758e+0 (9.27e-2) $-$\\
      \multirow{1}{*}{WFG6}&13&\hl{6.2270e+0 (7.35e-2)}&7.3923e+0 (1.95e-1) $-$&9.7253e+0 (4.17e-1) $-$&7.4216e+0 (1.69e-1) $-$&7.5724e+0 (1.74e-2) $-$&7.6153e+0 (3.15e-1) $-$\\
      \multirow{1}{*}{WFG6}&15&\hl{7.3575e+0 (1.20e-1)}&8.4147e+0 (1.37e-1) $-$&1.0518e+1 (7.38e-1) $-$&9.4706e+0 (2.82e-1) $-$&8.1808e+0 (3.52e-2) $-$&9.2472e+0 (3.54e-1) $-$\\
      \hline
      \multirow{1}{*}{WFG7}&5&\hl{1.1315e+0 (1.43e-2)}&1.1772e+0 (1.47e-3) $-$&1.1770e+0 (6.73e-4) $-$&1.3250e+0 (3.77e-2) $-$&1.1775e+0 (4.06e-4) $-$&1.1781e+0 (1.20e-3) $-$\\
      \multirow{1}{*}{WFG7}&10&\hl{4.3572e+0 (5.88e-2)}&4.7811e+0 (1.97e-2) $-$&4.8692e+0 (2.45e-1) $-$&5.0298e+0 (1.76e-1) $-$&4.7856e+0 (9.37e-3) $-$&4.6212e+0 (2.62e-2) $-$\\
      \multirow{1}{*}{WFG7}&13&\hl{6.0647e+0 (6.11e-2)}&7.6270e+0 (6.57e-2) $-$&7.3105e+0 (1.63e-1) $-$&7.4375e+0 (2.52e-1) $-$&7.3628e+0 (9.12e-2) $-$&6.8286e+0 (4.57e-1) $-$\\
      \multirow{1}{*}{WFG7}&15&\hl{7.3967e+0 (9.15e-2)}&8.3344e+0 (6.61e-2) $-$&8.8957e+0 (5.72e-1) $-$&9.5536e+0 (2.83e-1) $-$&8.1136e+0 (5.85e-2) $-$&7.7522e+0 (6.14e-1) $-$\\
      \hline
      \multirow{1}{*}{WFG8}&5&\hl{1.1214e+0 (1.02e-2)}&1.1579e+0 (4.79e-3) $-$&1.1489e+0 (3.16e-3) $-$&1.3597e+0 (4.00e-2) $-$&1.1476e+0 (2.38e-3) $-$&1.1667e+0 (1.28e-3) $-$\\
      \multirow{1}{*}{WFG8}&10&4.4215e+0 (8.44e-2)&4.7871e+0 (2.02e-2) $-$&4.7424e+0 (4.02e-1) $-$&5.3735e+0 (2.42e-1) $-$&4.5342e+0 (9.72e-2) $-$&\hl{4.3904e+0 (5.51e-2) $\approx$}\\
      \multirow{1}{*}{WFG8}&13&\hl{6.0768e+0 (6.30e-2)}&7.3980e+0 (2.36e-1) $-$&8.8582e+0 (5.78e-1) $-$&9.0387e+0 (2.62e-1) $-$&7.2280e+0 (2.02e-1) $-$&7.6203e+0 (3.76e-1) $-$\\
      \multirow{1}{*}{WFG8}&15&\hl{7.3036e+0 (1.35e-1)}&8.5820e+0 (4.16e-1) $-$&1.0303e+1 (8.03e-1) $-$&1.1449e+1 (4.26e-1) $-$&8.0169e+0 (3.26e-1) $-$&8.9501e+0 (3.12e-1) $-$\\
      \hline
      \multirow{1}{*}{WFG9}&5&\hl{1.1025e+0 (1.21e-2)}&1.1452e+0 (1.76e-3) $-$&1.1297e+0 (5.52e-3) $-$&1.2641e+0 (4.39e-2) $-$&1.1320e+0 (3.57e-3) $-$&1.1497e+0 (2.45e-3) $-$\\
      \multirow{1}{*}{WFG9}&10&4.4255e+0 (4.74e-2)&4.6449e+0 (1.78e-2) $-$&4.6187e+0 (1.95e-1) $-$&4.8168e+0 (1.41e-1) $-$&4.5239e+0 (3.21e-2) $-$&\hl{4.4169e+0 (6.29e-2) $\approx$}\\
      \multirow{1}{*}{WFG9}&13&\hl{5.9365e+0 (4.63e-2)}&7.2456e+0 (1.07e-1) $-$&7.3375e+0 (2.52e-1) $-$&7.4047e+0 (2.31e-1) $-$&7.1166e+0 (9.04e-2) $-$&6.6344e+0 (1.70e-1) $-$\\
      \multirow{1}{*}{WFG9}&15&\hl{6.9430e+0 (1.12e-1)}&7.7991e+0 (2.22e-1) $-$&8.1820e+0 (2.76e-1) $-$&9.0488e+0 (2.85e-1) $-$&7.5891e+0 (9.64e-2) $-$&7.4352e+0 (2.10e-1) $-$\\
      \hline
      \multirow{1}{*}{IDTLZ1}&5&6.7664e-2 (6.42e-4)&\hl{6.6289e-2 (1.19e-3) $+$}&1.4610e-1 (1.35e-2) $-$&7.2909e-2 (2.06e-3) $-$&1.6171e-1 (2.43e-2) $-$&1.4708e-1 (2.00e-2) $-$\\
      \multirow{1}{*}{IDTLZ1}&10&1.3290e-1 (6.29e-3)&\hl{1.1766e-1 (1.41e-3) $+$}&1.3394e-1 (3.91e-3) $\approx$&1.2977e-1 (3.17e-3) $\approx$&1.4561e-1 (4.74e-3) $-$&2.7719e-1 (4.11e-2) $-$\\
      \multirow{1}{*}{IDTLZ1}&13&1.5246e-1 (4.98e-3)&\hl{1.5037e-1 (7.13e-3) $\approx$}&1.6668e-1 (4.83e-3) $-$&1.6473e-1 (6.16e-3) $-$&1.8187e-1 (5.25e-3) $-$&3.0801e-1 (2.69e-2) $-$\\
      \multirow{1}{*}{IDTLZ1}&15&1.6496e-1 (2.34e-2)&1.6022e-1 (1.25e-2) $\approx$&\hl{1.5594e-1 (6.26e-3) $\approx$}&1.6676e-1 (4.60e-3) $-$&1.6697e-1 (5.42e-3) $-$&3.1018e-1 (3.14e-2) $-$\\
      \hline
      \multirow{1}{*}{IDTLZ2}&5&\hl{1.9447e-1 (2.87e-3)}&2.1510e-1 (4.02e-3) $-$&2.4555e-1 (1.00e-2) $-$&3.0992e-1 (1.32e-2) $-$&2.9606e-1 (2.27e-2) $-$&2.9736e-1 (5.06e-3) $-$\\
      \multirow{1}{*}{IDTLZ2}&10&\hl{4.1420e-1 (1.70e-3)}&5.4715e-1 (9.43e-3) $-$&5.2720e-1 (1.65e-2) $-$&5.8586e-1 (1.07e-2) $-$&6.4014e-1 (1.06e-2) $-$&6.6257e-1 (4.15e-3) $-$\\
      \multirow{1}{*}{IDTLZ2}&13&\hl{4.9414e-1 (1.63e-3)}&6.5560e-1 (1.01e-2) $-$&6.8806e-1 (1.99e-2) $-$&7.0846e-1 (1.07e-2) $-$&8.3026e-1 (9.42e-3) $-$&8.4693e-1 (1.65e-2) $-$\\
      \multirow{1}{*}{IDTLZ2}&15&\hl{5.3777e-1 (1.58e-3)}&6.6234e-1 (1.01e-2) $-$&6.9112e-1 (2.04e-2) $-$&7.1464e-1 (1.12e-2) $-$&8.3258e-1 (1.17e-2) $-$&8.4702e-1 (2.02e-2) $-$\\
      \hline
      \multicolumn{2}{c}{$+/-/\approx$}&&11/44/3&1/63/4&5/57/6&6/60/2&2/62/4\\
      \bottomrule
  \end{tabular}}
  \label{19IGD}
\end{table*}%
\begin{table*}[tbp]
  \renewcommand{\arraystretch}{1.2}
  \centering
  \caption{HV VALUES OBTAINED BY MAOEADPPS, \mbox{AR-MOEA~\cite{tian2017indicator}}, \mbox{NSGA-III~\cite{deb2014evolutionary}}, \mbox{RSEA~\cite{he2017radial}}, \mbox{$\theta $-DEA~\cite{yuan2016new}} AND RVEA~\cite{cheng2016reference} ON \mbox{DTLZ1-DTLZ6},  \mbox{IDTLZ1-IDTLZ2} AND \mbox{WFG1-WFG9} WITH 5, 10, 13 AND 15 OBJECTIVES. THE BEST RESULTS FOR EACH INSTANCE ARE HIGHLIGHTED IN GRAY.}
  \resizebox{0.96\textwidth}{!}{
    \begin{tabular}{cccccccc}
      \toprule
      Problem&$M$&MaOEADPPs&AR-MOEA&\mbox{NSGA-III}&RSEA&$\theta $-DEA&RVEA\\
      \midrule
      \multirow{1}{*}{DTLZ1}&5&9.7378e-1 (2.20e-3)&9.7496e-1 (1.47e-4) $+$&9.7486e-1 (1.59e-4) $+$&9.5810e-1 (4.99e-3) $-$&9.7489e-1 (1.67e-4) $+$&\hl{9.7503e-1 (1.99e-4) $+$}\\
      \multirow{1}{*}{DTLZ1}&10&\hl{9.9961e-1 (3.00e-5)}&9.9940e-1 (5.09e-4) $-$&9.2486e-1 (2.09e-1) $-$&9.9404e-1 (1.58e-3) $-$&9.9310e-1 (2.25e-2) $\approx$&9.9960e-1 (2.17e-5) $\approx$\\
      \multirow{1}{*}{DTLZ1}&13&\hl{9.9989e-1 (3.94e-5)}&9.9956e-1 (1.98e-4) $-$&9.5738e-1 (1.01e-1) $-$&9.8891e-1 (2.30e-2) $-$&9.9104e-1 (3.40e-2) $-$&9.9750e-1 (4.46e-3) $-$\\
      \multirow{1}{*}{DTLZ1}&15&\hl{9.9992e-1 (6.33e-5)}&9.9886e-1 (4.47e-3) $-$&9.0726e-1 (2.22e-1) $-$&9.9436e-1 (5.97e-3) $-$&9.9661e-1 (7.18e-3) $-$&9.9837e-1 (1.43e-3) $-$\\
      \hline
      \multirow{1}{*}{DTLZ2}&5&7.9424e-1 (1.02e-3)&7.9484e-1 (3.41e-4) $+$&7.9475e-1 (4.77e-4) $+$&7.5717e-1 (7.33e-3) $-$&\hl{7.9485e-1 (4.30e-4) $+$}&7.9482e-1 (3.99e-4) $+$\\
      \multirow{1}{*}{DTLZ2}&10&\hl{9.6894e-1 (8.75e-4)}&9.6737e-1 (2.60e-4) $-$&9.4576e-1 (4.07e-2) $-$&9.0762e-1 (1.06e-2) $-$&9.6762e-1 (2.09e-4) $-$&9.6772e-1 (2.01e-4) $-$\\
      \multirow{1}{*}{DTLZ2}&13&\hl{9.8442e-1 (4.03e-4)}&9.8038e-1 (4.19e-4) $-$&8.9812e-1 (4.43e-2) $-$&9.3433e-1 (1.15e-2) $-$&9.8070e-1 (1.98e-4) $-$&9.8094e-1 (1.76e-4) $-$\\
      \multirow{1}{*}{DTLZ2}&15&9.9030e-1 (6.45e-4)&\hl{9.9036e-1 (1.13e-3) $\approx$}&9.0708e-1 (5.16e-2) $-$&9.4728e-1 (8.41e-3) $-$&9.9032e-1 (2.08e-4) $\approx$&9.9029e-1 (2.72e-4) $\approx$\\
      \hline
      \multirow{1}{*}{DTLZ3}&5&7.8723e-1 (3.68e-3)&7.8650e-1 (6.35e-3) $\approx$&7.8388e-1 (1.01e-2) $\approx$&7.3230e-1 (3.90e-2) $-$&\hl{7.8872e-1 (4.31e-3) $+$}&7.5981e-1 (1.52e-1) $\approx$\\
      \multirow{1}{*}{DTLZ3}&10&\hl{9.5876e-1 (5.21e-3)}&8.6611e-1 (2.39e-1) $-$&7.1363e-2 (1.61e-1) $-$&4.8118e-1 (2.57e-1) $-$&8.0120e-1 (2.90e-1) $-$&9.5402e-1 (7.57e-3) $\approx$\\
      \multirow{1}{*}{DTLZ3}&13&\hl{9.7632e-1 (7.24e-3)}&9.5769e-1 (1.35e-2) $-$&0.0000e+0 (0.00e+0) $-$&2.5506e-1 (2.01e-1) $-$&8.8097e-1 (1.11e-1) $-$&9.7236e-1 (4.53e-3) $-$\\
      \multirow{1}{*}{DTLZ3}&15&\hl{9.7737e-1 (9.48e-3)}&9.6745e-1 (1.31e-2) $-$&0.0000e+0 (0.00e+0) $-$&7.7273e-2 (1.75e-1) $-$&8.8063e-1 (2.14e-1) $-$&9.3241e-1 (2.14e-1) $\approx$\\
      \hline
      \multirow{1}{*}{DTLZ4}&5&7.9474e-1 (1.10e-3)&7.9457e-1 (4.28e-4) $\approx$&7.7595e-1 (4.66e-2) $\approx$&7.5919e-1 (1.08e-2) $-$&\hl{7.9486e-1 (3.70e-4) $\approx$}&7.9139e-1 (1.77e-2) $\approx$\\
      \multirow{1}{*}{DTLZ4}&10&\hl{9.7029e-1 (2.33e-4)}&9.6817e-1 (2.49e-4) $-$&9.5205e-1 (3.31e-2) $-$&9.1721e-1 (6.86e-3) $-$&9.6780e-1 (1.76e-4) $-$&9.6786e-1 (1.98e-4) $-$\\
      \multirow{1}{*}{DTLZ4}&13&9.8535e-1 (3.16e-4)&\hl{9.8542e-1 (3.08e-4) $\approx$}&9.4227e-1 (3.59e-2) $-$&9.3596e-1 (1.26e-2) $-$&9.8123e-1 (1.42e-4) $-$&9.7981e-1 (3.06e-3) $-$\\
      \multirow{1}{*}{DTLZ4}&15&9.9042e-1 (2.50e-4)&\hl{9.9374e-1 (1.70e-4) $+$}&9.6761e-1 (2.42e-2) $\approx$&9.4240e-1 (1.17e-2) $-$&9.9129e-1 (1.35e-4) $+$&9.9122e-1 (1.10e-4) $+$\\
      \hline
      \multirow{1}{*}{DTLZ5}&5&1.1224e-1 (1.01e-3)&1.0879e-1 (3.91e-3) $-$&5.9147e-2 (4.96e-2) $-$&\hl{1.2217e-1 (1.38e-3) $+$}&9.9407e-2 (6.94e-3) $-$&1.0457e-1 (2.22e-3) $-$\\
      \multirow{1}{*}{DTLZ5}&10&\hl{9.3251e-2 (5.67e-4)}&8.2239e-2 (4.78e-3) $-$&9.1707e-3 (2.24e-2) $-$&7.9805e-2 (1.76e-2) $-$&9.2288e-2 (1.03e-3) $-$&9.0064e-2 (9.60e-4) $-$\\
      \multirow{1}{*}{DTLZ5}&13&\hl{9.2670e-2 (3.79e-4)}&9.1194e-2 (4.70e-4) $-$&0.0000e+0 (0.00e+0) $-$&7.3002e-2 (2.46e-2) $-$&9.1333e-2 (8.41e-4) $-$&9.1700e-2 (1.03e-3) $-$\\
      \multirow{1}{*}{DTLZ5}&15&\hl{9.2598e-2 (3.56e-4)}&8.4955e-2 (3.03e-3) $-$&2.3375e-3 (8.75e-3) $-$&7.4457e-2 (1.83e-2) $-$&9.0276e-2 (6.93e-4) $-$&9.0861e-2 (2.26e-4) $-$\\
      \hline
      \multirow{1}{*}{DTLZ6}&5&1.1097e-1 (5.85e-4)&1.0621e-1 (1.19e-2) $\approx$&5.4395e-2 (4.50e-2) $-$&\hl{1.1305e-1 (8.62e-3) $\approx$}&9.3731e-2 (4.38e-3) $-$&9.8356e-2 (1.78e-2) $-$\\
      \multirow{1}{*}{DTLZ6}&10&\hl{9.3511e-2 (5.97e-4)}&9.1861e-2 (1.15e-3) $-$&0.0000e+0 (0.00e+0) $-$&7.3266e-2 (3.63e-2) $\approx$&9.0904e-2 (1.06e-4) $-$&8.5324e-2 (2.36e-2) $-$\\
      \multirow{1}{*}{DTLZ6}&13&\hl{9.3287e-2 (4.74e-4)}&9.1130e-2 (3.66e-4) $-$&0.0000e+0 (0.00e+0) $-$&6.5443e-2 (3.14e-2) $-$&9.0944e-2 (1.75e-4) $-$&9.1080e-2 (3.18e-4) $-$\\
      \multirow{1}{*}{DTLZ6}&15&\hl{9.2705e-2 (3.80e-4)}&9.0382e-2 (2.64e-3) $-$&0.0000e+0 (0.00e+0) $-$&7.9738e-2 (1.81e-2) $-$&9.0926e-2 (1.88e-4) $-$&9.0854e-2 (1.93e-4) $-$\\
      \hline
      \multirow{1}{*}{WFG1}&5&\hl{9.9813e-1 (7.96e-5)}&9.9794e-1 (1.71e-4) $-$&9.8866e-1 (1.70e-2) $-$&9.9489e-1 (7.17e-4) $-$&9.9340e-1 (1.41e-3) $-$&9.9011e-1 (1.72e-2) $-$\\
      \multirow{1}{*}{WFG1}&10&\hl{9.9997e-1 (2.63e-5)}&9.0753e-1 (5.31e-2) $-$&8.0098e-1 (5.30e-2) $-$&9.8996e-1 (3.35e-2) $-$&9.9027e-1 (2.00e-2) $-$&9.5992e-1 (5.12e-2) $-$\\
      \multirow{1}{*}{WFG1}&13&\hl{9.9995e-1 (4.95e-5)}&9.8517e-1 (3.20e-2) $-$&9.9429e-1 (1.42e-2) $-$&9.9088e-1 (3.01e-2) $-$&9.9757e-1 (7.30e-4) $-$&9.5375e-1 (6.41e-2) $-$\\
      \multirow{1}{*}{WFG1}&15&\hl{9.9996e-1 (3.31e-5)}&9.9050e-1 (2.56e-2) $-$&9.1096e-1 (7.60e-2) $-$&9.9871e-1 (5.65e-4) $-$&9.9707e-1 (1.18e-3) $-$&9.7175e-1 (4.26e-2) $-$\\
      \hline
      \multirow{1}{*}{WFG2}&5&\hl{9.9676e-1 (5.47e-4)}&9.9569e-1 (9.11e-4) $-$&9.9498e-1 (5.90e-4) $-$&9.9289e-1 (1.29e-3) $-$&9.9396e-1 (8.43e-4) $-$&9.9182e-1 (1.93e-3) $-$\\
      \multirow{1}{*}{WFG2}&10&\hl{9.9844e-1 (9.09e-4)}&9.9375e-1 (1.41e-3) $-$&9.9605e-1 (1.65e-3) $-$&9.9723e-1 (1.04e-3) $-$&9.6648e-1 (1.48e-2) $-$&9.8085e-1 (3.74e-3) $-$\\
      \multirow{1}{*}{WFG2}&13&9.9730e-1 (1.70e-3)&9.9070e-1 (2.95e-3) $-$&9.9532e-1 (1.77e-3) $-$&\hl{9.9780e-1 (8.68e-4) $\approx$}&8.9781e-1 (7.20e-2) $-$&9.6366e-1 (6.63e-3) $-$\\
      \multirow{1}{*}{WFG2}&15&\hl{9.9731e-1 (1.49e-3)}&9.8876e-1 (4.27e-3) $-$&9.9597e-1 (1.60e-3) $-$&9.9701e-1 (1.04e-3) $\approx$&9.0098e-1 (7.81e-2) $-$&9.6956e-1 (1.34e-2) $-$\\
      \hline
      \multirow{1}{*}{WFG3}&5&2.1664e-1 (4.01e-3)&1.5023e-1 (1.47e-2) $-$&1.5494e-1 (1.56e-2) $-$&\hl{2.5602e-1 (1.46e-2) $+$}&2.0376e-1 (1.33e-2) $-$&1.4967e-1 (2.34e-2) $-$\\
      \multirow{1}{*}{WFG3}&10&\hl{7.7092e-2 (7.35e-3)}&0.0000e+0 (0.00e+0) $-$&4.9277e-3 (1.20e-2) $-$&5.2368e-2 (3.30e-2) $-$&1.7544e-2 (1.59e-2) $-$&0.0000e+0 (0.00e+0) $-$\\
      \multirow{1}{*}{WFG3}&13&\hl{8.0179e-3 (1.80e-2)}&0.0000e+0 (0.00e+0) $-$&0.0000e+0 (0.00e+0) $-$&3.3755e-3 (8.64e-3) $\approx$&0.0000e+0 (0.00e+0) $-$&0.0000e+0 (0.00e+0) $-$\\
      \multirow{1}{*}{WFG3}&15&0.0000e+0 (0.00e+0)&\hl{0.0000e+0 (0.00e+0) $\approx$}&0.0000e+0 (0.00e+0) $\approx$&0.0000e+0 (0.00e+0) $\approx$&0.0000e+0 (0.00e+0) $\approx$&0.0000e+0 (0.00e+0) $\approx$\\
      \hline
      \multirow{1}{*}{WFG4}&5&\hl{7.9406e-1 (9.88e-4)}&7.8971e-1 (1.26e-3) $-$&7.8551e-1 (1.80e-2) $-$&7.5527e-1 (4.66e-3) $-$&7.9027e-1 (8.15e-4) $-$&7.8938e-1 (1.39e-3) $-$\\
      \multirow{1}{*}{WFG4}&10&9.3894e-1 (3.85e-3) $-$&9.3843e-1 (1.92e-2) $-$&9.1207e-1 (7.77e-3) $-$&9.4511e-1 (2.49e-3) $-$&9.3761e-1 (6.24e-3) $-$&\hl{9.6299e-1 (2.09e-3)}\\
      \multirow{1}{*}{WFG4}&13&\hl{9.8185e-1 (1.31e-3)}&9.6216e-1 (7.88e-3) $-$&8.9704e-1 (3.33e-2) $-$&9.4034e-1 (6.52e-3) $-$&9.4720e-1 (4.97e-3) $-$&9.4144e-1 (1.18e-2) $-$\\
      \multirow{1}{*}{WFG4}&15&\hl{9.8589e-1 (1.71e-3)}&9.6713e-1 (7.52e-3) $-$&8.9832e-1 (2.43e-2) $-$&9.4605e-1 (1.00e-2) $-$&9.4328e-1 (5.55e-3) $-$&9.4118e-1 (9.70e-3) $-$\\
      \hline
      \multirow{1}{*}{WFG5}&5&7.4349e-1 (1.03e-3)&7.4362e-1 (3.84e-4) $\approx$&7.4351e-1 (4.79e-4) $\approx$&7.0393e-1 (6.75e-3) $-$&\hl{7.4368e-1 (4.38e-4) $\approx$}&7.4343e-1 (5.06e-4) $\approx$\\
      \multirow{1}{*}{WFG5}&10&\hl{9.0232e-1 (6.81e-4)}&8.9193e-1 (1.71e-3) $-$&8.9624e-1 (5.54e-4) $-$&8.4815e-1 (7.22e-3) $-$&8.9687e-1 (9.29e-4) $-$&8.9529e-1 (1.36e-3) $-$\\
      \multirow{1}{*}{WFG5}&13&\hl{9.0802e-1 (1.42e-3)}&8.9374e-1 (2.99e-3) $-$&8.7228e-1 (3.33e-2) $-$&8.6783e-1 (8.18e-3) $-$&8.9277e-1 (3.19e-3) $-$&9.0288e-1 (1.09e-3) $-$\\
      \multirow{1}{*}{WFG5}&15&\hl{9.1073e-1 (1.49e-3)}&8.9410e-1 (5.19e-3) $-$&8.8312e-1 (1.40e-2) $-$&8.8085e-1 (4.09e-3) $-$&8.7855e-1 (7.20e-3) $-$&9.0087e-1 (3.07e-3) $-$\\
      \hline
      \multirow{1}{*}{WFG6}&5&\hl{7.3073e-1 (2.14e-2)}&7.2615e-1 (1.37e-2) $\approx$&7.2260e-1 (1.56e-2) $\approx$&6.9498e-1 (1.67e-2) $-$&7.2442e-1 (1.73e-2) $\approx$&7.2691e-1 (1.72e-2) $\approx$\\
      \multirow{1}{*}{WFG6}&10&\hl{9.0289e-1 (2.29e-2)}&8.6723e-1 (1.86e-2) $-$&8.4887e-1 (5.12e-2) $-$&8.3793e-1 (1.82e-2) $-$&8.6258e-1 (1.79e-2) $-$&8.5110e-1 (2.86e-2) $-$\\
      \multirow{1}{*}{WFG6}&13&\hl{9.2044e-1 (2.17e-2)}&8.7257e-1 (2.03e-2) $-$&7.5981e-1 (4.96e-2) $-$&8.4890e-1 (1.34e-2) $-$&8.7475e-1 (1.86e-2) $-$&7.2856e-1 (6.96e-2) $-$\\
      \multirow{1}{*}{WFG6}&15&\hl{9.2806e-1 (2.13e-2)}&8.6916e-1 (2.72e-2) $-$&7.9551e-1 (3.07e-2) $-$&8.5393e-1 (2.67e-2) $-$&8.7260e-1 (2.09e-2) $-$&7.5043e-1 (9.36e-2) $-$\\
      \hline
      \multirow{1}{*}{WFG7}&5&\hl{7.9381e-1 (1.47e-3)}&7.8654e-1 (1.41e-3) $-$&7.9064e-1 (7.00e-4) $-$&7.5739e-1 (7.19e-3) $-$&7.9187e-1 (5.64e-4) $-$&7.8881e-1 (8.34e-4) $-$\\
      \multirow{1}{*}{WFG7}&10&\hl{9.6923e-1 (7.20e-4)}&9.4508e-1 (4.33e-3) $-$&9.4262e-1 (1.68e-2) $-$&9.1110e-1 (1.06e-2) $-$&9.5681e-1 (1.59e-3) $-$&9.3922e-1 (3.34e-3) $-$\\
      \multirow{1}{*}{WFG7}&13&\hl{9.8607e-1 (7.14e-4)}&9.5981e-1 (4.34e-3) $-$&9.4566e-1 (1.66e-2) $-$&9.4329e-1 (8.75e-3) $-$&9.5461e-1 (5.06e-3) $-$&9.4989e-1 (9.60e-3) $-$\\
      \multirow{1}{*}{WFG7}&15&\hl{9.9054e-1 (1.10e-3)}&9.3857e-1 (5.57e-3) $-$&9.2664e-1 (2.62e-2) $-$&9.4403e-1 (9.76e-3) $-$&9.4651e-1 (6.39e-3) $-$&9.3374e-1 (1.98e-2) $-$\\
      \hline
      \multirow{1}{*}{WFG8}&5&\hl{6.8730e-1 (3.42e-3)}&6.8016e-1 (1.27e-3) $-$&6.8168e-1 (1.85e-3) $-$&6.2452e-1 (6.20e-3) $-$&6.8115e-1 (2.10e-3) $-$&6.7139e-1 (1.74e-3) $-$\\
      \multirow{1}{*}{WFG8}&10&\hl{9.0212e-1 (1.28e-2)}&8.9089e-1 (1.95e-2) $\approx$&8.4058e-1 (3.56e-2) $-$&8.1471e-1 (8.55e-3) $-$&8.6471e-1 (1.50e-2) $-$&7.7255e-1 (9.14e-2) $-$\\
      \multirow{1}{*}{WFG8}&13&\hl{9.2754e-1 (1.30e-2)}&8.8618e-1 (4.53e-2) $-$&6.6590e-1 (9.63e-2) $-$&8.2079e-1 (1.63e-2) $-$&8.2825e-1 (1.81e-2) $-$&6.3181e-1 (8.63e-2) $-$\\
      \multirow{1}{*}{WFG8}&15&\hl{9.3813e-1 (1.24e-2)}&8.4650e-1 (6.47e-2) $-$&5.9513e-1 (1.37e-1) $-$&8.2920e-1 (2.02e-2) $-$&8.5913e-1 (1.47e-2) $-$&7.2041e-1 (1.27e-1) $-$\\
      \hline
      \multirow{1}{*}{WFG9}&5&\hl{7.6337e-1 (3.17e-2)}&7.2808e-1 (5.22e-3) $-$&7.3382e-1 (2.63e-2) $-$&7.2241e-1 (8.19e-3) $-$&7.4880e-1 (3.56e-3) $-$&7.4185e-1 (2.76e-2) $-$\\
      \multirow{1}{*}{WFG9}&10&\hl{9.0524e-1 (8.94e-3)}&8.3013e-1 (2.21e-2) $-$&8.2624e-1 (5.37e-2) $-$&8.6052e-1 (4.15e-3) $-$&8.8447e-1 (7.03e-3) $-$&8.3133e-1 (4.38e-2) $-$\\
      \multirow{1}{*}{WFG9}&13&\hl{9.0450e-1 (1.44e-2)}&7.9056e-1 (4.51e-2) $-$&8.3025e-1 (6.04e-2) $-$&8.8783e-1 (8.99e-3) $-$&8.3956e-1 (2.78e-2) $-$&8.2223e-1 (4.23e-2) $-$\\
      \multirow{1}{*}{WFG9}&15&\hl{9.0535e-1 (1.59e-2)}&7.7997e-1 (4.57e-2) $-$&8.3843e-1 (7.51e-2) $-$&8.9839e-1 (8.75e-3) $\approx$&8.4051e-1 (1.94e-2) $-$&7.7658e-1 (5.10e-2) $-$\\
      \hline
      \multirow{1}{*}{IDTLZ1}&5&9.9458e-3 (1.39e-4)&\hl{1.0170e-2 (3.33e-4) $+$}&3.6347e-3 (4.26e-4) $-$&9.7749e-3 (2.17e-4) $-$&3.2677e-3 (9.27e-4) $-$&2.7020e-3 (8.05e-4) $-$\\
      \multirow{1}{*}{IDTLZ1}&10&1.3057e-7 (3.45e-7)&\hl{7.3241e-7 (9.61e-7) $+$}&4.9091e-7 (5.11e-8) $+$&5.4925e-7 (5.51e-8) $+$&4.7776e-7 (5.08e-8) $+$&1.2525e-8 (6.90e-9) $-$\\
      \multirow{1}{*}{IDTLZ1}&13&0.0000e+0 (0.00e+0)&0.0000e+0 (0.00e+0) $\approx$&5.1457e-10 (1.46e-10) $+$&\hl{6.0675e-10 (1.27e-10) $+$}&4.3014e-10 (9.47e-11) $+$&3.9345e-12 (2.09e-12) $+$\\
      \multirow{1}{*}{IDTLZ1}&15&0.0000e+0 (0.00e+0)&2.2193e-13 (8.30e-13) $\approx$&6.1093e-12 (1.72e-12) $+$&5.3803e-12 (1.94e-12) $+$&\hl{6.1315e-12 (1.22e-12) $+$}&3.5611e-14 (1.61e-14) $+$\\
      \hline
      \multirow{1}{*}{IDTLZ2}&5&\hl{1.0438e-1 (1.15e-3)}&9.4386e-2 (2.39e-3) $-$&6.3268e-2 (6.46e-3) $-$&1.0333e-1 (1.95e-3) $\approx$&7.6298e-2 (5.34e-3) $-$&6.2500e-2 (2.16e-3) $-$\\
      \multirow{1}{*}{IDTLZ2}&10&9.0750e-5 (1.12e-5)&1.8225e-4 (8.72e-6) $+$&1.1825e-4 (5.07e-5) $\approx$&\hl{3.9341e-4 (1.09e-5) $+$}&3.3035e-4 (1.50e-5) $+$&1.6171e-4 (1.25e-5) $+$\\
      \multirow{1}{*}{IDTLZ2}&13&6.9981e-7 (5.17e-7)&3.7964e-6 (4.68e-7) $+$&2.5984e-6 (8.90e-7) $+$&\hl{7.3562e-6 (2.55e-7) $+$}&3.2051e-6 (2.62e-7) $+$&1.7201e-6 (6.06e-7) $+$\\
      \multirow{1}{*}{IDTLZ2}&15&5.6855e-8 (2.13e-7)&2.4490e-7 (3.37e-8) $+$&1.9433e-7 (6.45e-8) $+$&\hl{4.3134e-7 (1.59e-8) $+$}&1.7783e-7 (1.85e-8) $+$&9.6859e-8 (3.48e-8) $+$\\
      \hline
      \multicolumn{2}{c}{$+/-/\approx$}&&15/48/5&7/41/10&7/53/8&9/53/6&8/51/9\\
      \bottomrule
  \end{tabular}}
  \label{19HV}
\end{table*}

Table~\ref{19IGD} shows that MaOEADPPs achieves the best IGD values on 45 out of 68 benchmark instances of \mbox{DTLZ1-DTLZ6}, \mbox{IDTLZ1-IDTLZ2} and \mbox{WFG1-WFG9} with~5, 10, 13, and~15 objectives.
\mbox{AR-MOEA} achieves the best results on~10, \mbox{NSGA-III} on~1, RSEA on~5, \mbox{$\theta$-DEA} on~4, and RVEA on~3 of these instances. From Table~\ref{19HV}, we find that MaOEADPPs achieves the best HV values on~47 of the same 68~instances, whereas \mbox{AR-MOEA} is best on~6, RSEA on~8, \mbox{$\theta$-DEA} on~5, and RVEA on 2~problems.%
  \begin{table*}[tbp]
  \renewcommand{\arraystretch}{1.2}
  \centering
  \caption{IGD VALUES OBTAINED BY MAOEADPPS, \mbox{AR-MOEA~\cite{tian2017indicator}}, \mbox{NSGA-III~\cite{deb2014evolutionary}}, RSEA~\cite{he2017radial}, \mbox{$\theta $-DEA~\cite{yuan2016new}} AND RVEA~\cite{cheng2016reference} ON \mbox{MAF1-MAF7} WITH 5, 10, 13 AND 15~OBJECTIVES. THE BEST RESULTS FOR EACH INSTANCE ARE HIGHLIGHTED IN GRAY.}
  \resizebox{\textwidth}{!}{
    \begin{tabular}{cccccccc}
      \toprule
      Problem&$M$&MaOEADPPs&AR-MOEA&\mbox{NSGA-III}&RSEA&$\theta $-DEA&RVEA\\
      \midrule
      \multirow{1}{*}{MaF1}&5&\hl{1.3381e-1 (6.64e-4)}&1.3444e-1 (1.82e-3) $-$&2.9054e-1 (2.72e-2) $-$&1.4759e-1 (3.28e-3) $-$&3.1258e-1 (3.91e-2) $-$&2.9663e-1 (3.39e-2) $-$\\
      \multirow{1}{*}{MaF1}&10&2.6866e-1 (3.33e-3)&\hl{2.4207e-1 (1.18e-3) $+$}&2.7454e-1 (1.21e-2) $\approx$&2.5208e-1 (5.85e-3) $+$&2.9709e-1 (7.88e-3) $-$&6.1872e-1 (8.00e-2) $-$\\
      \multirow{1}{*}{MaF1}&13&3.1280e-1 (6.57e-3)&\hl{2.9581e-1 (4.50e-3) $+$}&3.3621e-1 (7.54e-3) $-$&3.1624e-1 (7.75e-3) $\approx$&3.6579e-1 (1.23e-2) $-$&6.8961e-1 (4.33e-2) $-$\\
      \multirow{1}{*}{MaF1}&15&\hl{3.0834e-1 (6.76e-3)}&3.2145e-1 (5.81e-3) $-$&3.1575e-1 (6.70e-3) $-$&3.1962e-1 (1.07e-2) $-$&3.3156e-1 (5.77e-3) $-$&6.5207e-1 (4.54e-2) $-$\\
      \hline
      \multirow{1}{*}{MaF2}&5&1.1366e-1 (1.40e-3)&\hl{1.1256e-1 (1.52e-3) $+$}&1.2973e-1 (2.73e-3) $-$&1.2556e-1 (7.35e-3) $-$&1.3990e-1 (4.77e-3) $-$&1.3057e-1 (1.89e-3) $-$\\
      \multirow{1}{*}{MaF2}&10&\hl{1.7499e-1 (2.96e-3)}&2.0318e-1 (8.73e-3) $-$&2.2546e-1 (3.44e-2) $-$&3.2254e-1 (2.10e-2) $-$&2.1180e-1 (1.27e-2) $-$&4.0219e-1 (2.03e-1) $-$\\
      \multirow{1}{*}{MaF2}&13&\hl{1.7796e-1 (2.39e-3)}&2.3099e-1 (8.22e-3) $-$&2.4010e-1 (3.12e-2) $-$&3.1551e-1 (2.18e-2) $-$&2.5570e-1 (3.03e-2) $-$&4.9529e-1 (9.93e-2) $-$\\
      \multirow{1}{*}{MaF2}&15&\hl{1.8511e-1 (1.47e-3)}&2.3170e-1 (1.14e-2) $-$&2.1123e-1 (1.72e-2) $-$&3.0455e-1 (1.54e-2) $-$&2.8104e-1 (1.71e-2) $-$&4.7995e-1 (1.50e-1) $-$\\
      \hline
      \multirow{1}{*}{MaF3}&5&8.2383e-2 (1.57e-3)&8.6819e-2 (2.33e-3) $-$&1.0050e-1 (4.67e-2) $-$&1.1766e-1 (1.04e-1) $\approx$&1.0298e-1 (1.45e-3) $-$&\hl{7.5637e-2 (6.76e-3) $+$}\\
      \multirow{1}{*}{MaF3}&10&\hl{9.7108e-2 (7.17e-3)}&3.6116e+0 (1.33e+1) $\approx$&1.3875e+4 (2.54e+4) $-$&6.4126e+2 (7.34e+2) $-$&2.7136e-1 (1.85e-1) $-$&2.0124e-1 (3.51e-1) $-$\\
      \multirow{1}{*}{MaF3}&13&1.1530e-1 (6.26e-3)&4.7458e+0 (1.16e+1) $\approx$&5.4994e+5 (1.54e+6) $-$&2.1263e+3 (2.60e+3) $-$&9.5223e-1 (1.63e+0) $-$&\hl{1.1164e-1 (8.34e-3) $\approx$}\\
      \multirow{1}{*}{MaF3}&15&\hl{1.0752e-1 (4.79e-3)}&1.8651e+1 (4.65e+1) $\approx$&5.1341e+4 (1.52e+5) $-$&1.7411e+3 (1.43e+3) $-$&5.2407e+0 (1.92e+1) $-$&1.4855e-1 (1.44e-1) $-$\\
      \hline
      \multirow{1}{*}{MaF4}&5&\hl{2.0758e+0 (6.12e-2)}&2.5465e+0 (7.06e-2) $-$&2.9373e+0 (1.53e-1) $-$&2.9801e+0 (1.33e-1) $-$&3.9221e+0 (8.80e-1) $-$&4.1637e+0 (1.10e+0) $-$\\
      \multirow{1}{*}{MaF4}&10&\hl{5.9697e+1 (1.17e+1)}&9.6482e+1 (4.21e+0) $-$&9.5999e+1 (5.65e+0) $-$&1.0492e+2 (1.05e+1) $-$&1.1165e+2 (8.28e+0) $-$&1.8399e+2 (5.28e+1) $-$\\
      \multirow{1}{*}{MaF4}&13&\hl{3.5266e+2 (6.22e+1)}&9.9651e+2 (1.26e+2) $-$&8.9650e+2 (9.62e+1) $-$&8.8779e+2 (9.34e+1) $-$&1.0234e+3 (1.15e+2) $-$&1.8603e+3 (3.77e+2) $-$\\
      \multirow{1}{*}{MaF4}&15&\hl{1.4325e+3 (3.22e+2)}&4.1427e+3 (5.14e+2) $-$&3.5664e+3 (3.19e+2) $-$&3.3671e+3 (3.35e+2) $-$&4.4368e+3 (3.63e+2) $-$&8.1299e+3 (1.60e+3) $-$\\
      \hline
      \multirow{1}{*}{MaF5}&5&\hl{2.1053e+0 (3.14e-2)}&2.3914e+0 (2.79e-2) $-$&2.5317e+0 (4.18e-1) $-$&2.5003e+0 (1.82e-1) $-$&2.5812e+0 (8.26e-1) $-$&2.6886e+0 (9.32e-1) $-$\\
      \multirow{1}{*}{MaF5}&10&\hl{6.2725e+1 (3.14e+0)}&9.8822e+1 (4.42e+0) $-$&8.7509e+1 (8.66e-1) $-$&8.7670e+1 (1.48e+1) $-$&8.8046e+1 (1.14e+0) $-$&9.3621e+1 (7.40e+0) $-$\\
      \multirow{1}{*}{MaF5}&13&\hl{3.7946e+2 (3.11e+1)}&1.0252e+3 (3.18e+1) $-$&7.9584e+2 (1.85e+1) $-$&5.5773e+2 (1.07e+2) $-$&7.9882e+2 (1.29e+1) $-$&8.5434e+2 (8.66e+1) $-$\\
      \multirow{1}{*}{MaF5}&15&\hl{1.4607e+3 (1.24e+2)}&3.6857e+3 (1.37e+2) $-$&2.4772e+3 (5.97e+1) $-$&2.3899e+3 (3.95e+2) $-$&2.4579e+3 (5.17e+1) $-$&3.0355e+3 (3.48e+2) $-$\\
      \hline
      \multirow{1}{*}{MaF6}&5&1.1935e-2 (2.75e-3)&\hl{3.4567e-3 (4.60e-5) $+$}&9.4379e-2 (1.13e-1) $-$&3.8166e-1 (9.18e-2) $-$&1.2034e-1 (9.33e-3) $-$&1.3493e-1 (1.45e-1) $-$\\
      \multirow{1}{*}{MaF6}&10&\hl{7.3264e-3 (2.46e-3)}&1.3999e-1 (1.80e-1) $\approx$&5.9221e-1 (3.15e-1) $-$&4.2057e-1 (2.24e-1) $-$&1.6571e-1 (1.59e-1) $-$&1.3661e-1 (1.74e-2) $-$\\
      \multirow{1}{*}{MaF6}&13&\hl{7.8533e-3 (3.04e-3)}&2.6975e-1 (1.89e-1) $-$&5.6373e-1 (2.22e-1) $-$&4.5514e-1 (1.28e-1) $-$&2.9188e-1 (1.72e-1) $-$&4.3864e-1 (2.80e-1) $-$\\
      \multirow{1}{*}{MaF6}&15&\hl{6.5037e-3 (1.10e-3)}&3.0348e-1 (1.03e-1) $-$&4.3582e+0 (9.36e+0) $-$&6.3933e-1 (2.66e-1) $-$&4.7849e-1 (1.26e-1) $-$&3.8822e-1 (2.75e-1) $-$\\
      \hline
      \multirow{1}{*}{MaF7}&5&\hl{3.0435e-1 (3.17e-2)}&3.3360e-1 (1.04e-2) $-$&3.4907e-1 (2.66e-2) $-$&4.8673e-1 (1.21e-1) $-$&4.2825e-1 (3.63e-2) $-$&5.5657e-1 (1.17e-2) $-$\\
      \multirow{1}{*}{MaF7}&10&\hl{9.9091e-1 (2.02e-2)}&1.5334e+0 (1.16e-1) $-$&1.2252e+0 (9.98e-2) $-$&2.2322e+0 (5.93e-1) $-$&1.1269e+0 (1.35e-1) $-$&2.2404e+0 (4.55e-1) $-$\\
      \multirow{1}{*}{MaF7}&13&\hl{1.3975e+0 (6.78e-2)}&3.7647e+0 (7.32e-1) $-$&1.6700e+0 (1.95e-1) $-$&4.5661e+0 (9.68e-1) $-$&1.9826e+0 (5.59e-1) $-$&2.4885e+0 (2.83e-1) $-$\\
      \multirow{1}{*}{MaF7}&15&\hl{1.7859e+0 (7.82e-2)}&3.1638e+0 (7.25e-1) $-$&3.5061e+0 (5.56e-1) $-$&8.2792e+0 (1.03e+0) $-$&3.6564e+0 (5.68e-1) $-$&2.7712e+0 (4.58e-1) $-$\\
      \hline
      \multicolumn{2}{c}{$+/-/\approx$}&&4/20/4&0/27/1&1/25/2&0/28/0&1/26/1\\
      \bottomrule
  \end{tabular}}
  \label{MAF}
\end{table*}

Table~\ref{MAF} shows the IGD values reached on \mbox{MaF1-MaF7} with~5, 10, 13 and 15~objectives.
Here, MaOEADPPs achieves the best values on~22 of the 28~instances, while \mbox{AR-MOEA} is best on~4 and RVEA on~2 of the instances.

The instances we selected include different types of MaOPs. \mbox{DTLZ1-DTLZ4} as well as WFG1, 4, 5, and~7 are separable. WFG2, 3, 6, 8, and~9 are nonseparable. WFG4 and WFG9 as well as DTLZ1 and~DTLZ3 and MaF3, 4, and~7 are multimodal, the others are unimodal. \mbox{WFG1-WFG2} and MaF3 are convex. MaF6 and WFG3 are degenerate. \mbox{WFG4-WFG9} are concave and MaF7 and WFG3 are disconnected. We therefore confirmed that MaOEADPPs is effective on a wide variety of MaOPs.
\subsection{Effectiveness of DPPs-Selection}%
\begin{figure*}[tb]
  \centering

  \begin{minipage}{0.3\linewidth}

    \centerline{\includegraphics[width=4.7cm]{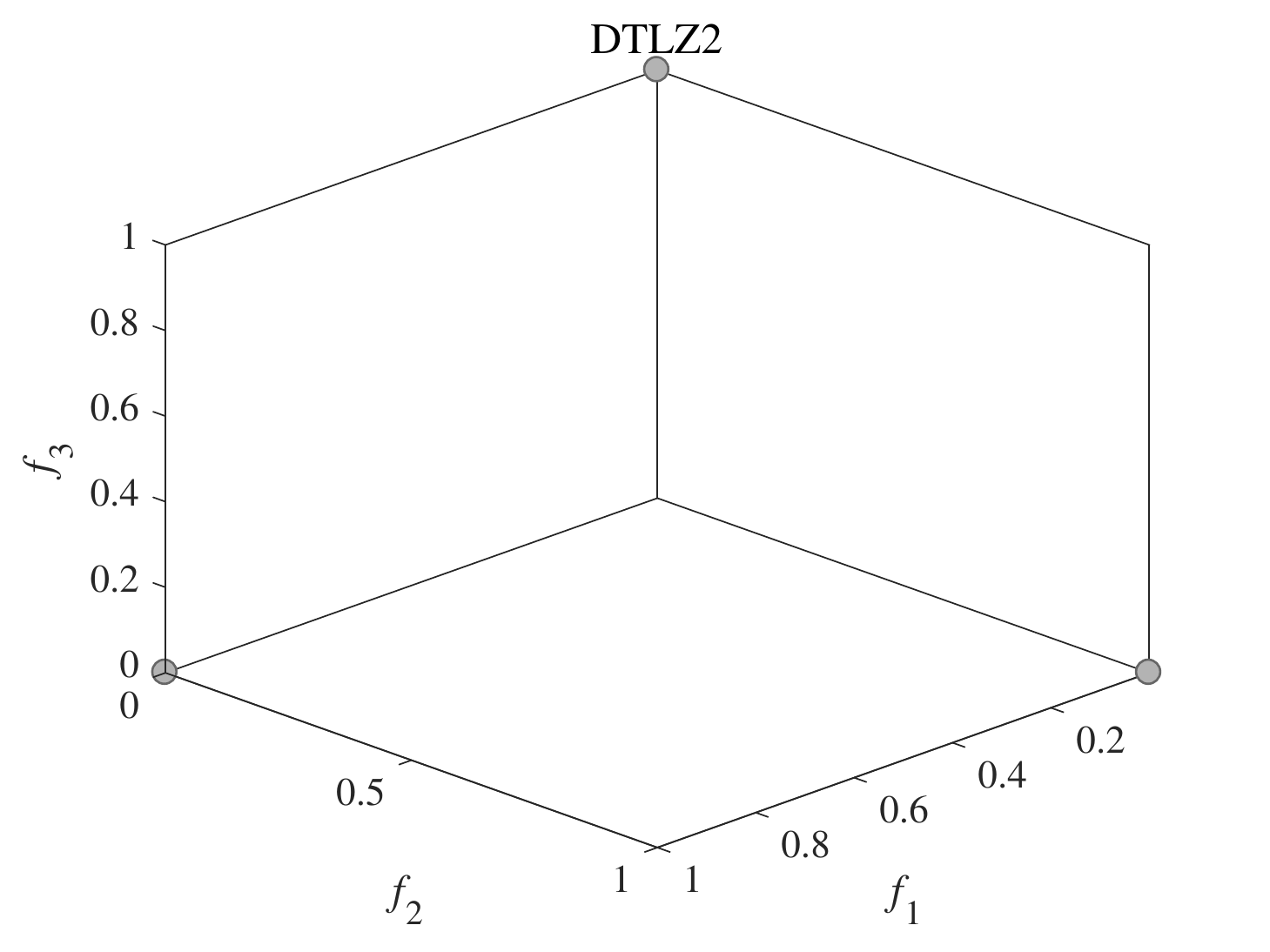}}

    \centerline{(a) }

  \end{minipage}
  \begin{minipage}{.3\linewidth}

    \centerline{\includegraphics[width=4.7cm]{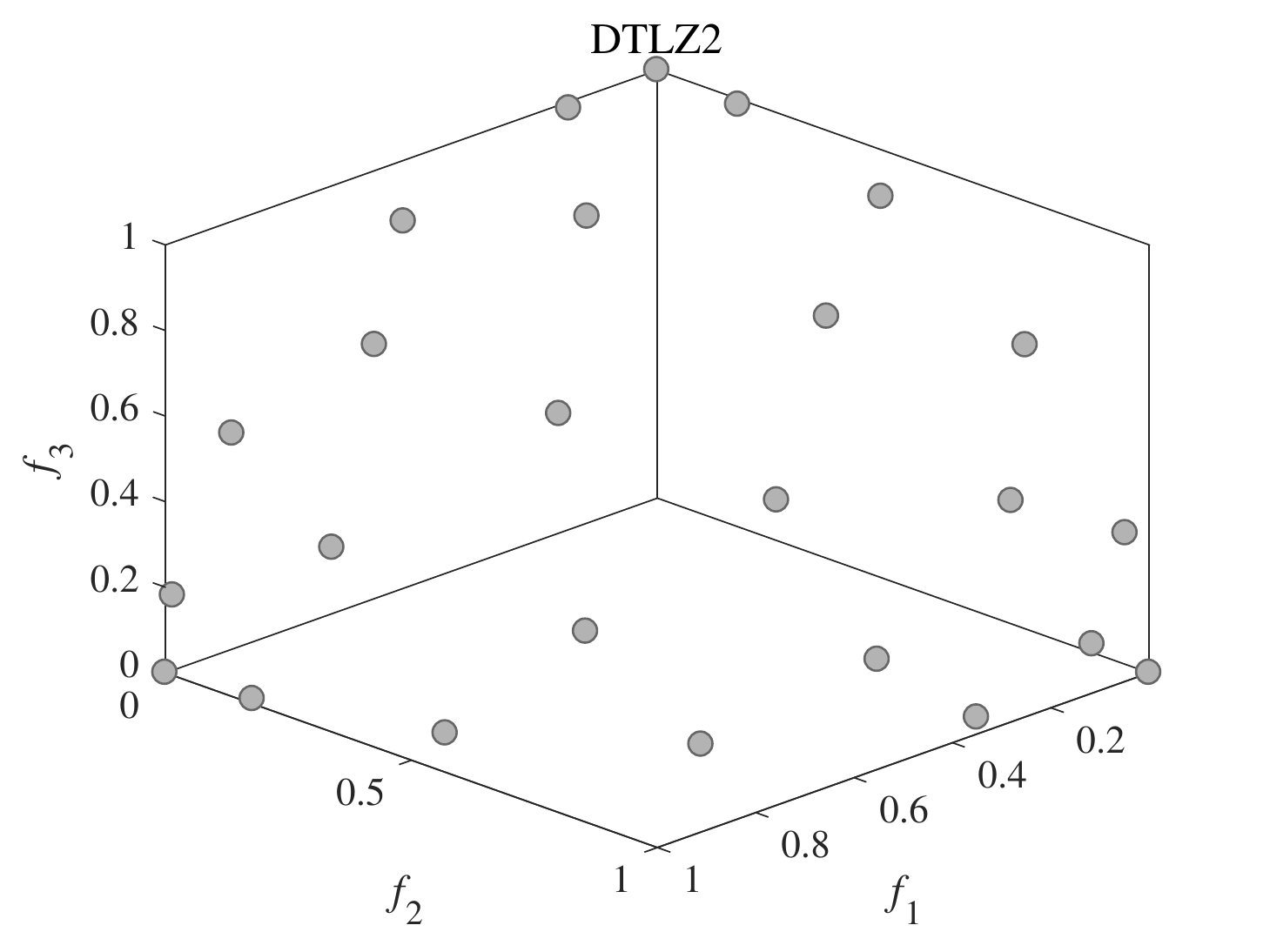}}

    \centerline{(b) }

  \end{minipage}
  \begin{minipage}{0.3\linewidth}

    \centerline{\includegraphics[width=4.7cm]{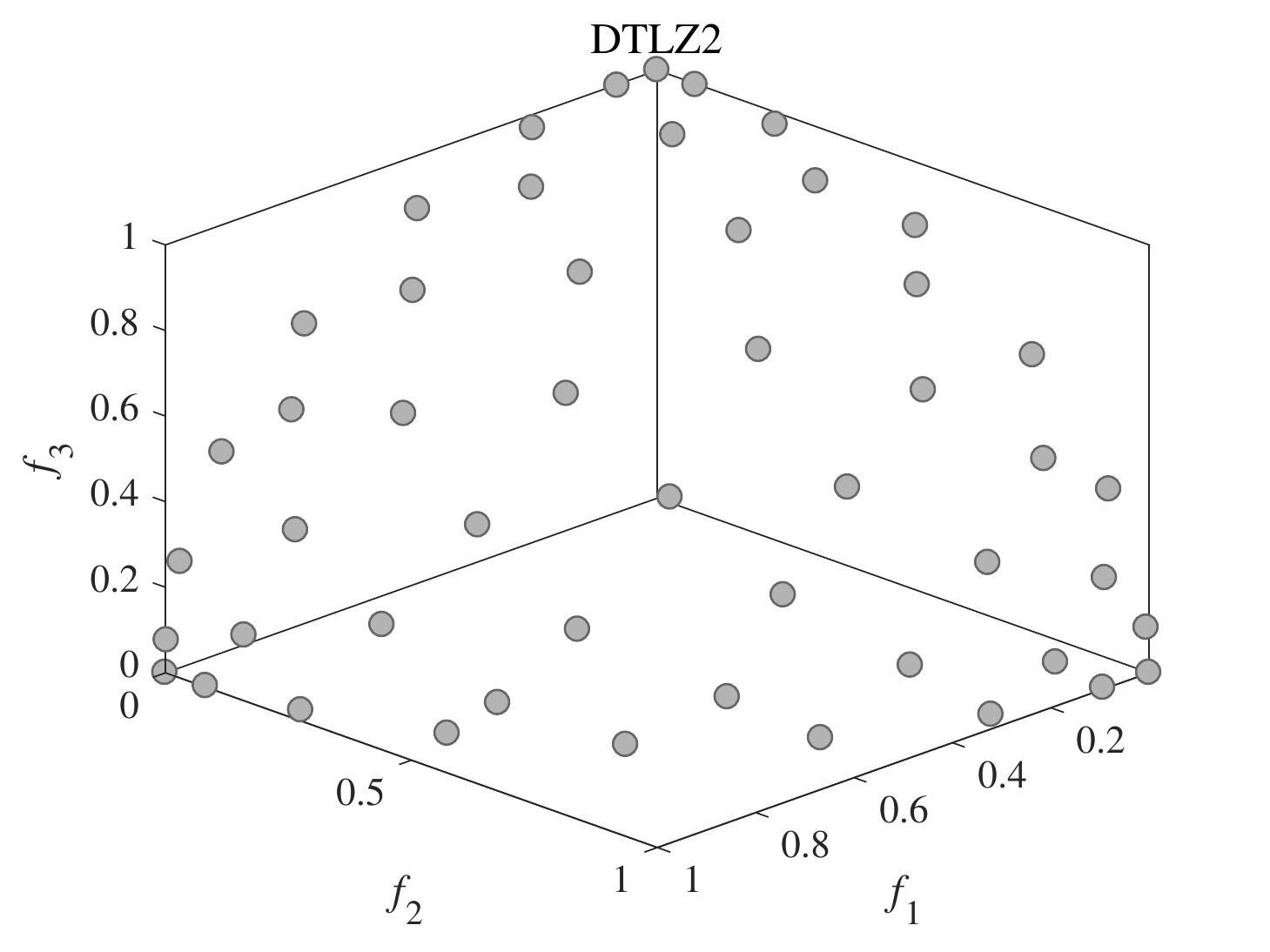}}

    \centerline{(c)}

  \end{minipage}

  \begin{minipage}{0.3\linewidth}

    \centerline{\includegraphics[width=4.7cm]{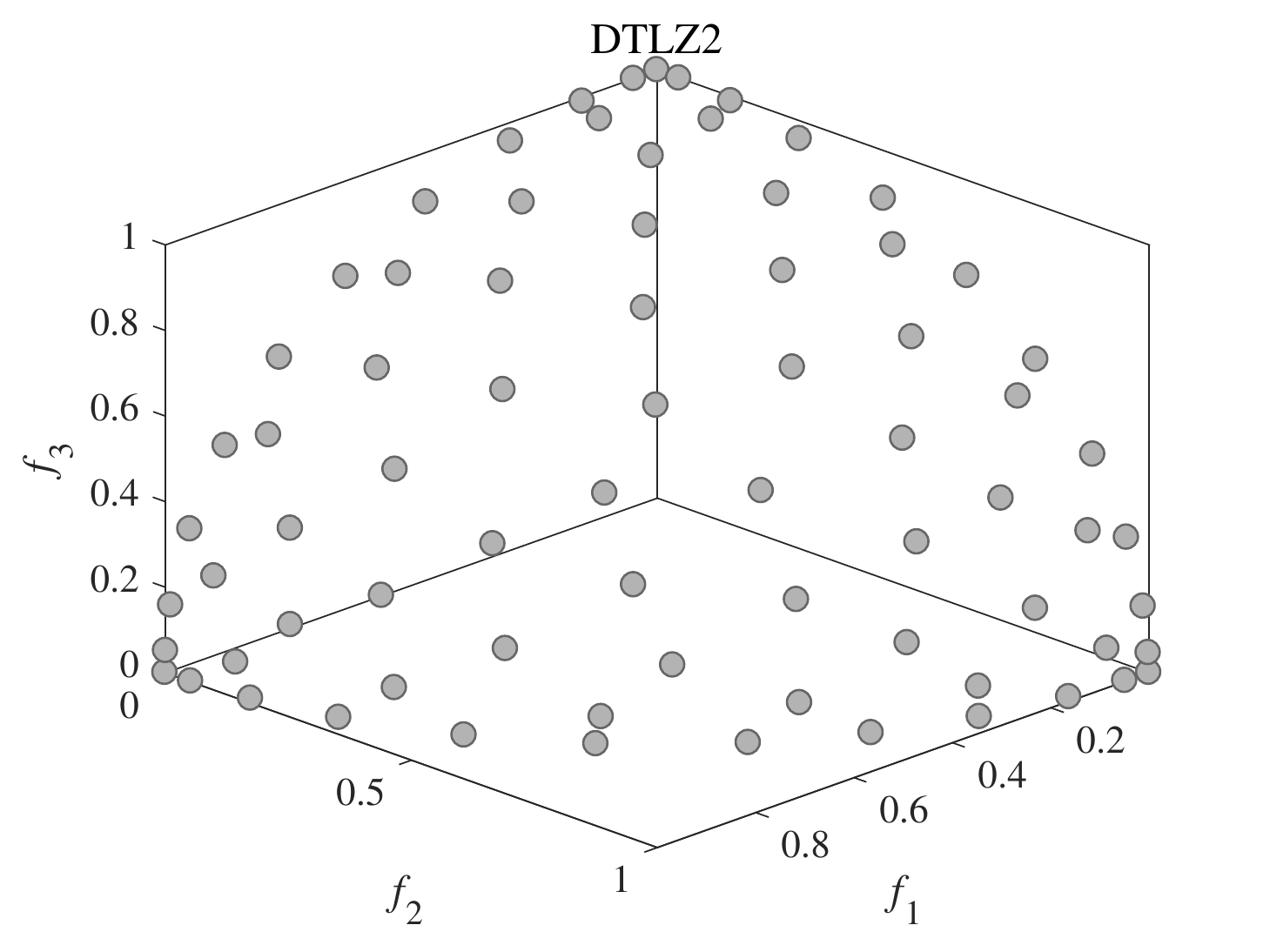}}

    \centerline{(d)}

  \end{minipage}
  \begin{minipage}{0.3\linewidth}

    \centerline{\includegraphics[width=4.7cm]{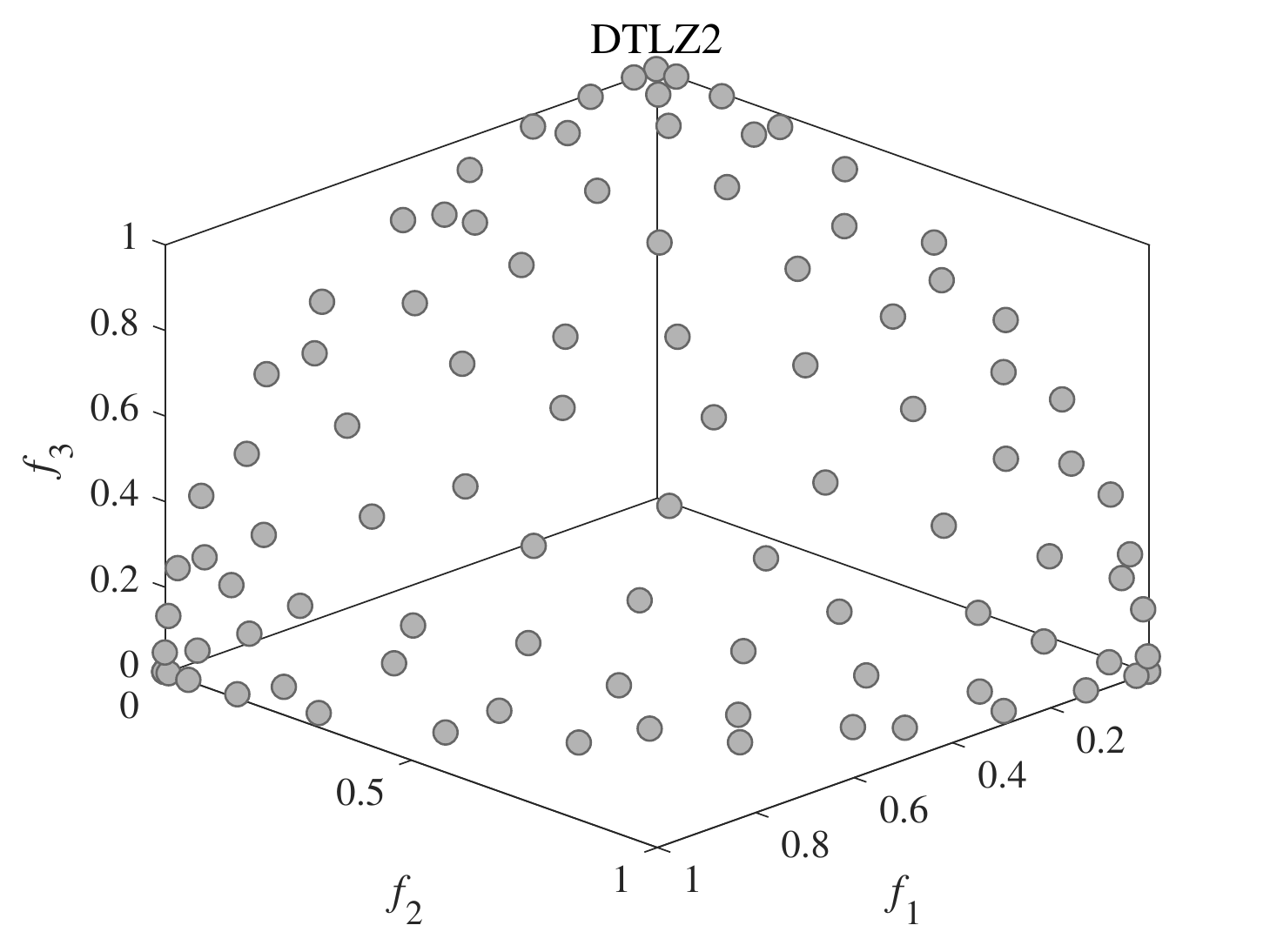}}

    \centerline{(e)}

  \end{minipage}
  \begin{minipage}{0.3\linewidth}

    \centerline{\includegraphics[width=4.7cm]{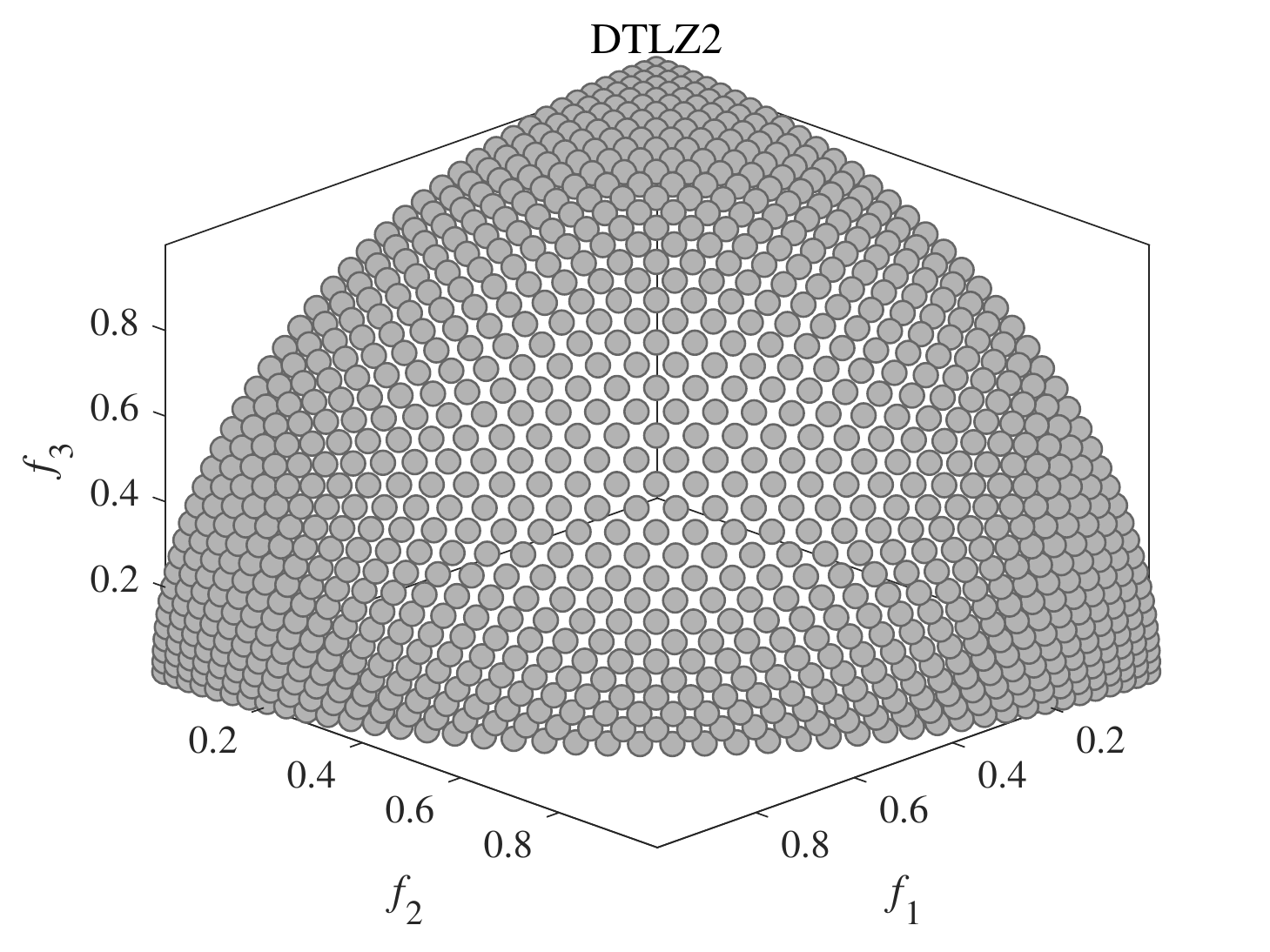}}

    \centerline{(f)}

  \end{minipage}
\caption{Example of the process of choosing solutions from a PF by DPPs-Selection: (a)~the first selected 3 points; (b)~the first selected 25 points; (c)~the first selected 50 points; (d)~the first selected 75 points; (e)~the first selected 100 points; (f)~the true PF. }%
\label{fig:res_1}%
\end{figure*}%
We now investigate the effectiveness of DPPs-Selection and whether it can maintain diversity.
In order to show the diversity of the solutions chosen by DPPs-Selection, we carry out experiments on DTLZ2.
We set the population sizes to~3, 25, 50, 75, and~100 and the number of function evaluations to~100'000.
\mbox{Figure~\ref{fig:res_1} (a)-(e)} shows the results of DPPs-Selection selecting~3, 25, 50, 75, and~100 solutions from a true PF, which is shown in Figure~\ref{fig:res_1}~(f). Regardless of the population size, MaOEADPPs maintains the diversity of the population. It can describe the distribution of true PF well with just a limited number of solutions.

To further evaluate the effectiveness of the DPPs-Selection, we replace it with \mbox{$k$-DPPs} and uniform sampling. \mbox{$k$-DPPs} selects eigenvectors based on the probability calculated by Equation~(\ref{hh}) while DPPs-Selection selects the eigenvectors with $k$~largest eigenvalues.

We apply the modified algorithm to each instance independently for 30~times. We compare the experimental results of \mbox{$k$-DPPs}, DPPs-Selection and uniform sampling on DTLZ1-DTLZ6 with~5, 10, 15~objectives. The population size is set to~126 for 5~objectives, 230 for 10~objectives, and 240 for 15~objectives. The number of function evaluations is set to~100'000.

The IGD values are shown in Table~\ref{KDPPs_SS}. The experimental results demonstrate that DPPs-Selection outperforms the other methods on~14 of the 18~instances. This clearly indicates that is effective in selecting good offspring.%
\begin{table*}[tbp]
  \renewcommand{\arraystretch}{1.2}
  \centering
  \caption{IGD VALUES OF DPPS-SELECTION, \mbox{$k$-DPPS}, AND UNIFORM SAMPLING ON \mbox{DTLZ1-DTLZ6} WITH 5, 10 AND 15~OBJECTIVES. THE BEST RESULTS FOR EACH INSTANCE ARE HIGHLIGHTED IN GRAY.}
  \begin{tabular}{cccccc}
    \toprule
    Problem&$M$&$D$&DPPs-Selection&$k$-DPPs&Uniform Sampling\\
    \midrule
    \multirow{1}{*}{DTLZ1}&5&9&\hl{6.3302e-2 (2.01e-3) $+$}&6.7296e-2 (1.50e-3) $+$&1.0132e-1 (3.60e-3)\\
    \multirow{1}{*}{DTLZ1}&10&14&1.1160e-1 (7.24e-4) $+$&\hl{1.1097e-1 (1.41e-3) $+$}&1.3398e-1 (3.13e-3)\\
    \multirow{1}{*}{DTLZ1}&15&19&1.3144e-1 (2.25e-3) $+$&\hl{1.2933e-1 (3.66e-3) $+$}&1.4420e-1 (3.78e-3)\\
    \hline
    \multirow{1}{*}{DTLZ2}&5&14&\hl{1.9256e-1 (9.47e-4) $+$}&2.1073e-1 (2.27e-3) $+$&4.7890e-1 (1.76e-2)\\
    \multirow{1}{*}{DTLZ2}&10&19&\hl{4.1889e-1 (3.08e-3) $+$}&4.2137e-1 (2.30e-3) $+$&6.4167e-1 (1.04e-2)\\
    \multirow{1}{*}{DTLZ2}&15&24&\hl{5.2576e-1 (2.53e-3) $+$}&5.2848e-1 (2.48e-3) $+$&7.1932e-1 (1.42e-2)\\
    \hline
    \multirow{1}{*}{DTLZ3}&5&14&\hl{1.9863e-1 (3.76e-3) $+$}&2.1074e-1 (3.26e-3) $+$&4.9174e-1 (1.57e-2)\\
    \multirow{1}{*}{DTLZ3}&10&19&\hl{4.2238e-1 (7.54e-3) $+$}&4.3082e-1 (1.58e-2) $+$&6.3344e-1 (1.65e-2)\\
    \multirow{1}{*}{DTLZ3}&15&24&\hl{5.2696e-1 (5.94e-3) $+$}&5.4254e-1 (1.76e-2) $+$&6.9256e-1 (1.95e-2)\\
    \hline
    \multirow{1}{*}{DTLZ4}&5&14&\hl{1.9286e-1 (9.25e-4) $+$}&2.0636e-1 (2.29e-3) $+$&4.0326e-1 (5.62e-2)\\
    \multirow{1}{*}{DTLZ4}&10&19&4.2837e-1 (2.71e-3) $+$&\hl{4.2613e-1 (1.70e-3) $+$}&6.0528e-1 (1.35e-2)\\
    \multirow{1}{*}{DTLZ4}&15&24&\hl{5.2782e-1 (9.74e-4) $+$}&5.3930e-1 (2.47e-3) $+$&6.9463e-1 (1.45e-2)\\
    \hline
    \multirow{1}{*}{DTLZ5}&5&14&\hl{5.4818e-2 (3.13e-3) $+$}&6.1737e-2 (1.11e-2) $+$&2.0254e-1 (4.49e-2)\\
    \multirow{1}{*}{DTLZ5}&10&19&\hl{8.6127e-2 (9.47e-3) $+$}&9.4108e-2 (1.49e-2) $+$&1.8816e-1 (3.92e-2)\\
    \multirow{1}{*}{DTLZ5}&15&24&1.3073e-1 (1.76e-2) $+$&\hl{1.0559e-1 (1.33e-2) $+$}&1.7968e-1 (2.09e-2)\\
    \hline
    \multirow{1}{*}{DTLZ6}&5&14&\hl{5.7469e-2 (3.00e-3) $+$}&8.1320e-2 (1.26e-2) $+$&2.3739e-1 (3.75e-2)\\
    \multirow{1}{*}{DTLZ6}&10&19&\hl{7.8330e-2 (9.20e-3) $+$}&1.3341e-1 (3.87e-2) $+$&2.0934e-1 (4.00e-2)\\
    \multirow{1}{*}{DTLZ6}&15&24&\hl{1.1411e-1 (1.48e-2) $+$}&1.6549e-1 (4.05e-2) $\approx$&2.0659e-1 (3.55e-2)\\
    \hline
    \multicolumn{3}{c}{$+/-/\approx$}&18/0/0&17/0/1&\\
    \bottomrule
  \end{tabular}
  \label{KDPPs_SS}
\end{table*}%
\begin{figure*}[tbp]
  \centering

  \begin{minipage}{0.3\linewidth}

    \centerline{\includegraphics[width=3.5cm]{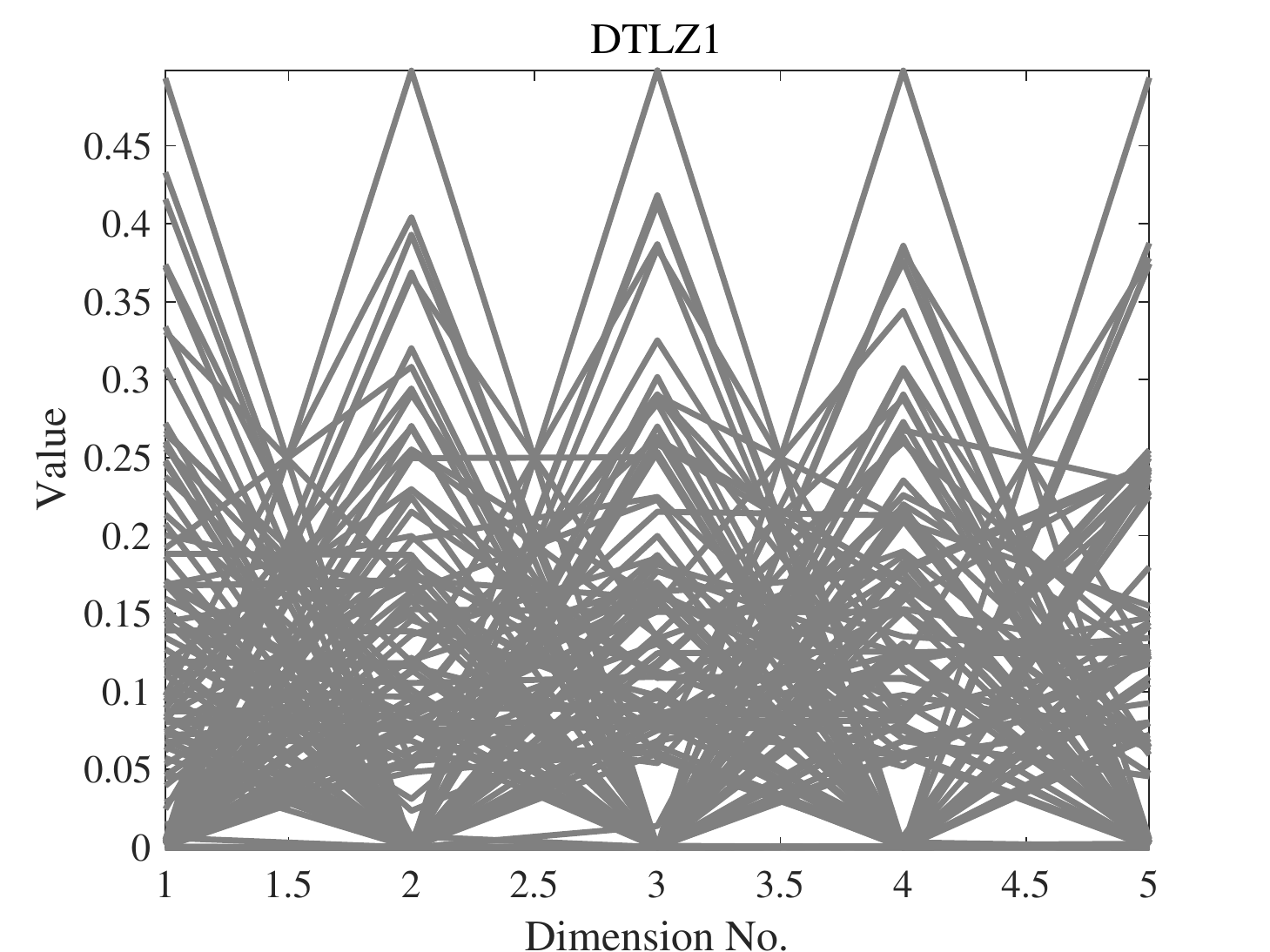}}

    \centerline{(a) DPPs-Selection}

  \end{minipage}
  \begin{minipage}{.3\linewidth}

    \centerline{\includegraphics[width=3.5cm]{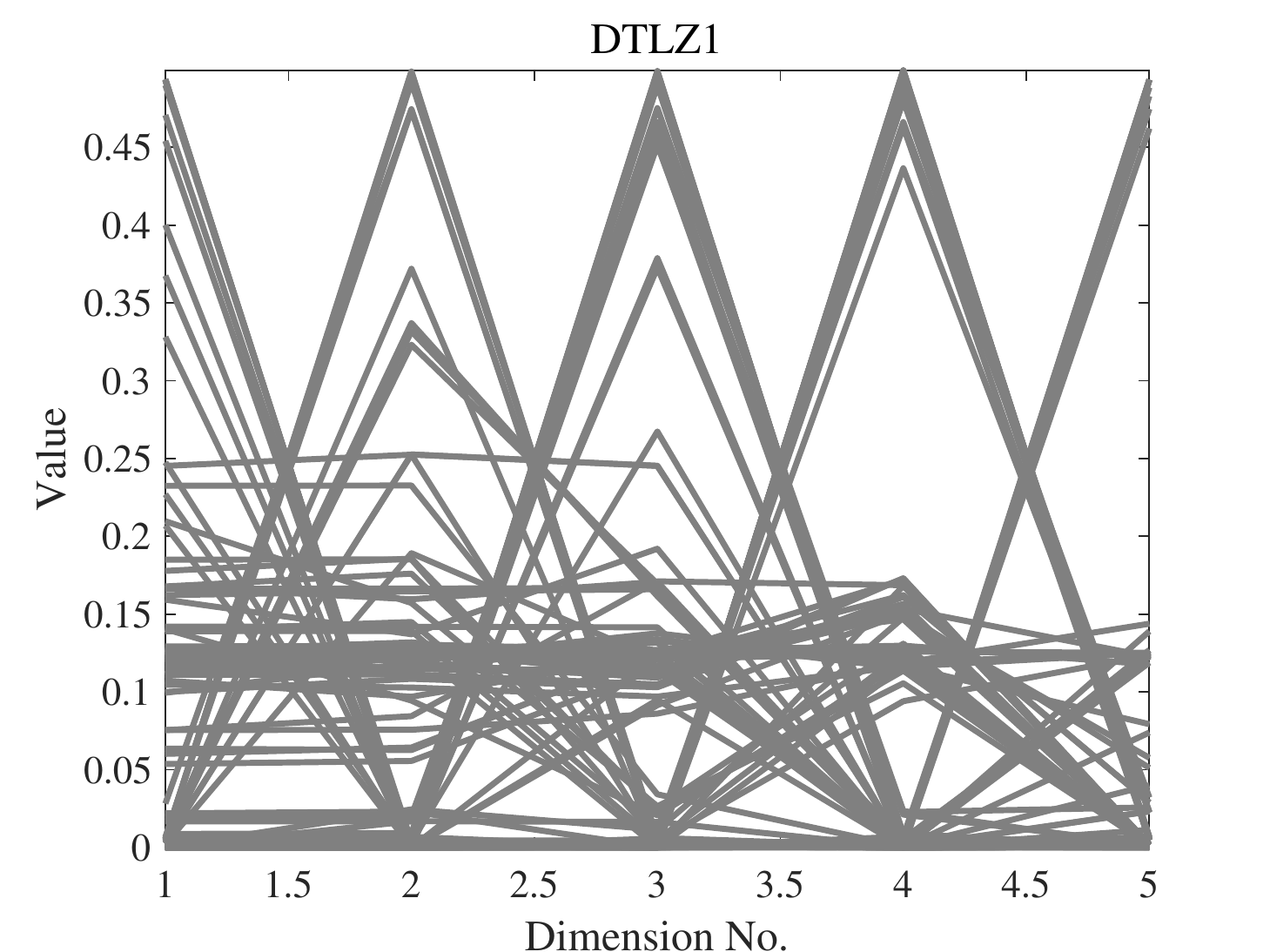}}

    \centerline{(b) Uniform sampling }

  \end{minipage}
  \begin{minipage}{0.3\linewidth}

    \centerline{\includegraphics[width=3.5cm]{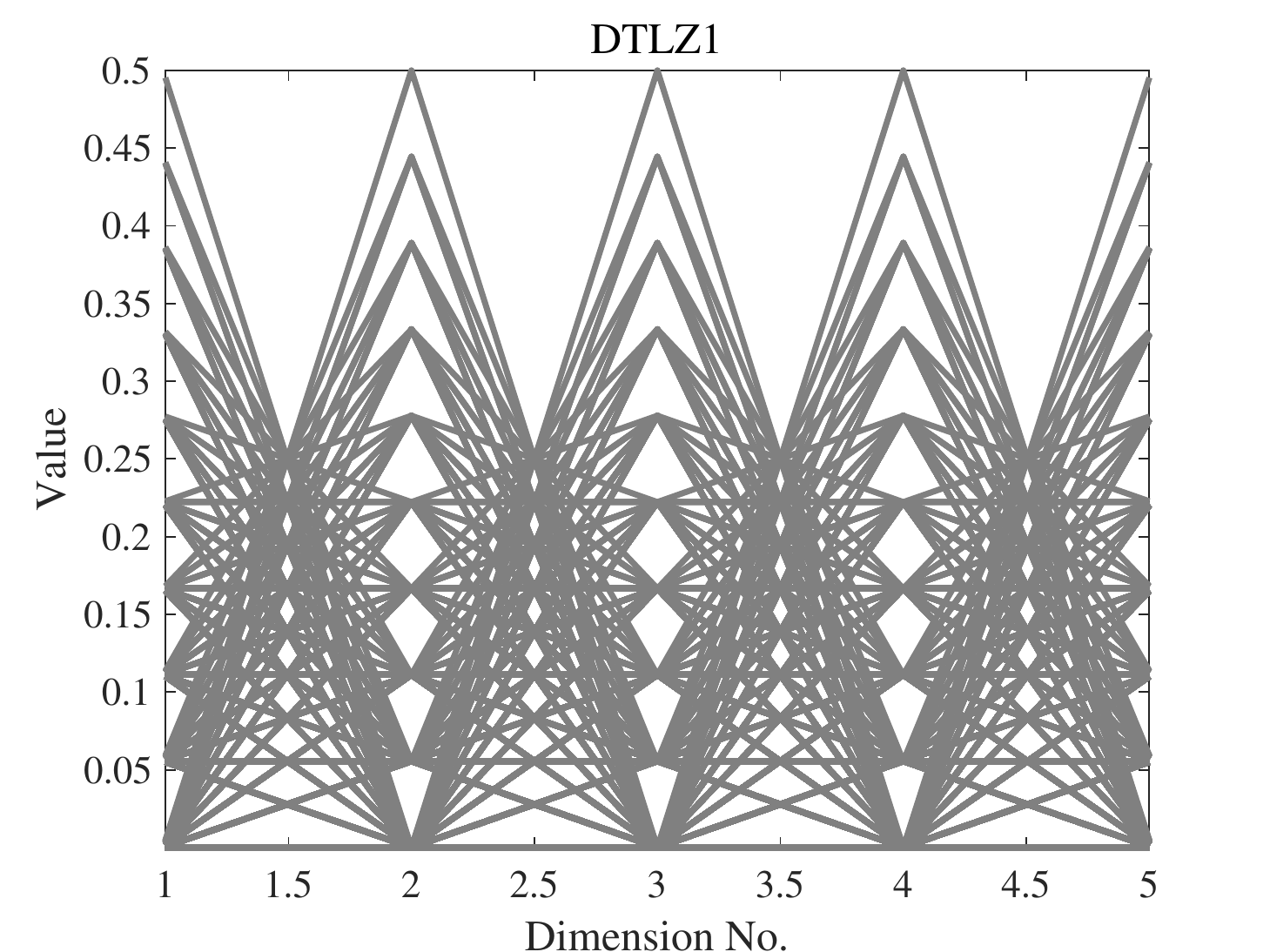}}

    \centerline{(c) True PF}

  \end{minipage}

  \begin{minipage}{0.3\linewidth}

    \centerline{\includegraphics[width=3.5cm]{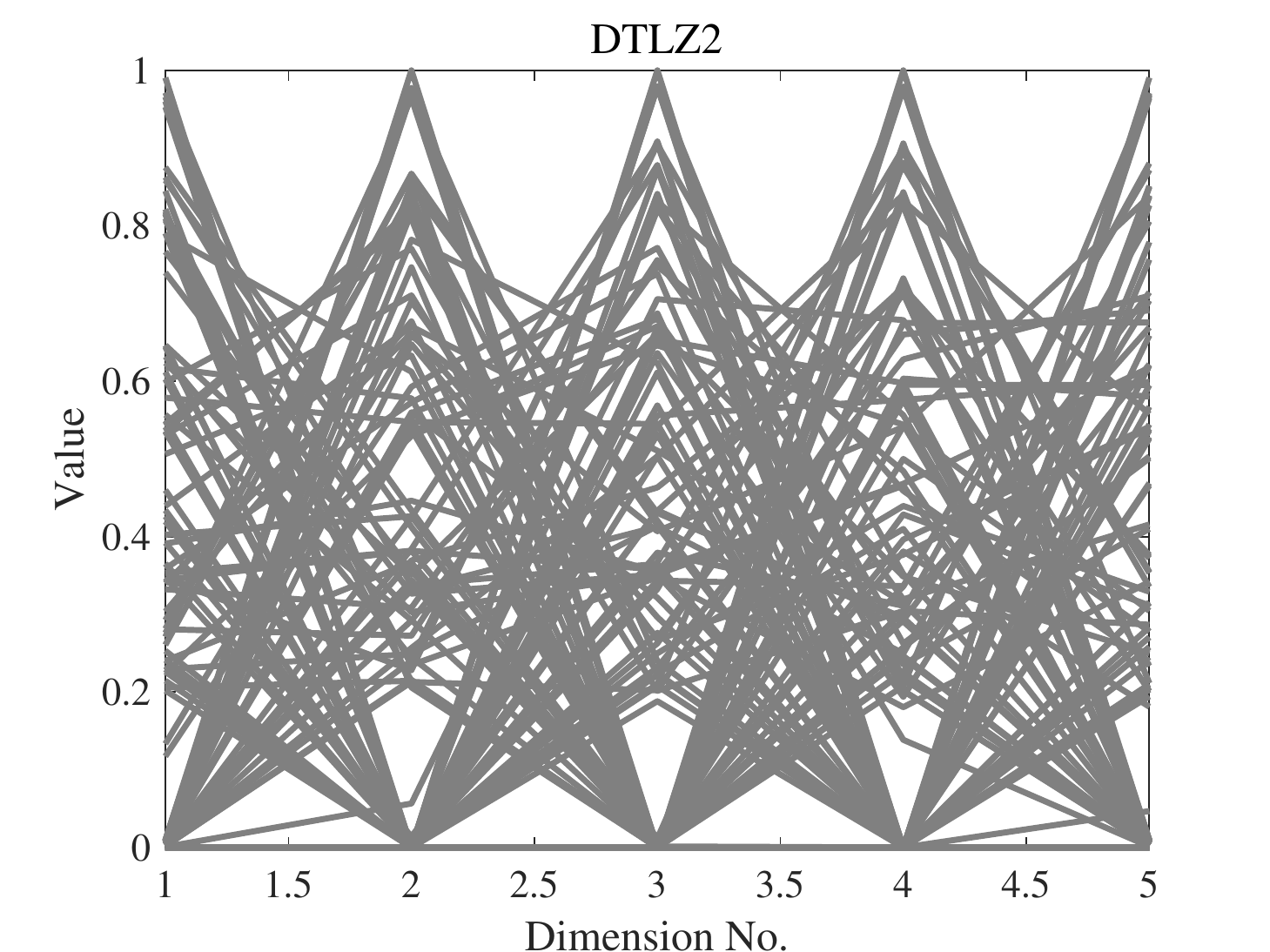}}

    \centerline{(d) DPPs-Selection}

  \end{minipage}
  \begin{minipage}{0.3\linewidth}

    \centerline{\includegraphics[width=3.5cm]{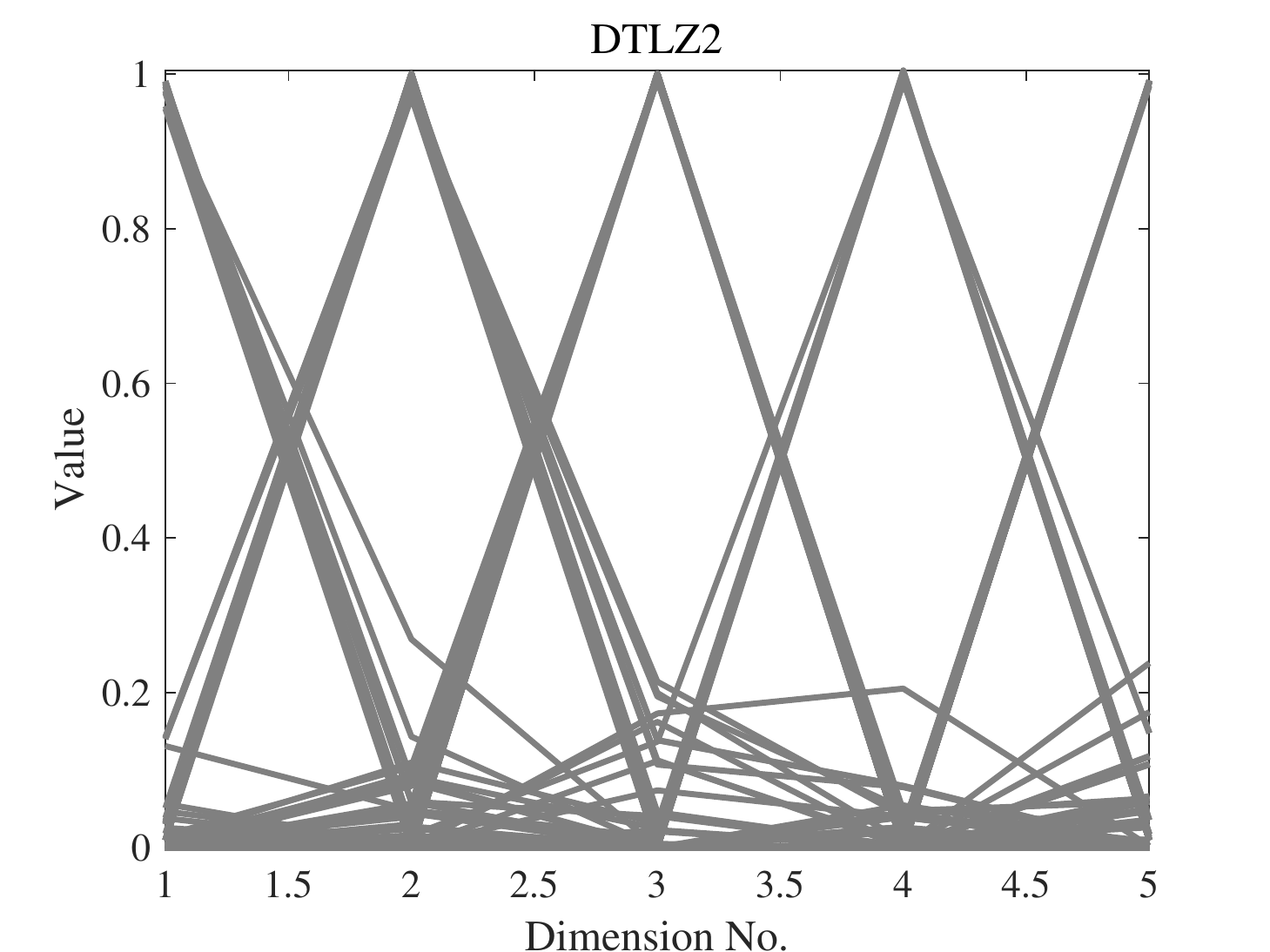}}

    \centerline{(e)  Uniform sampling}

  \end{minipage}
  \begin{minipage}{0.3\linewidth}

    \centerline{\includegraphics[width=3.5cm]{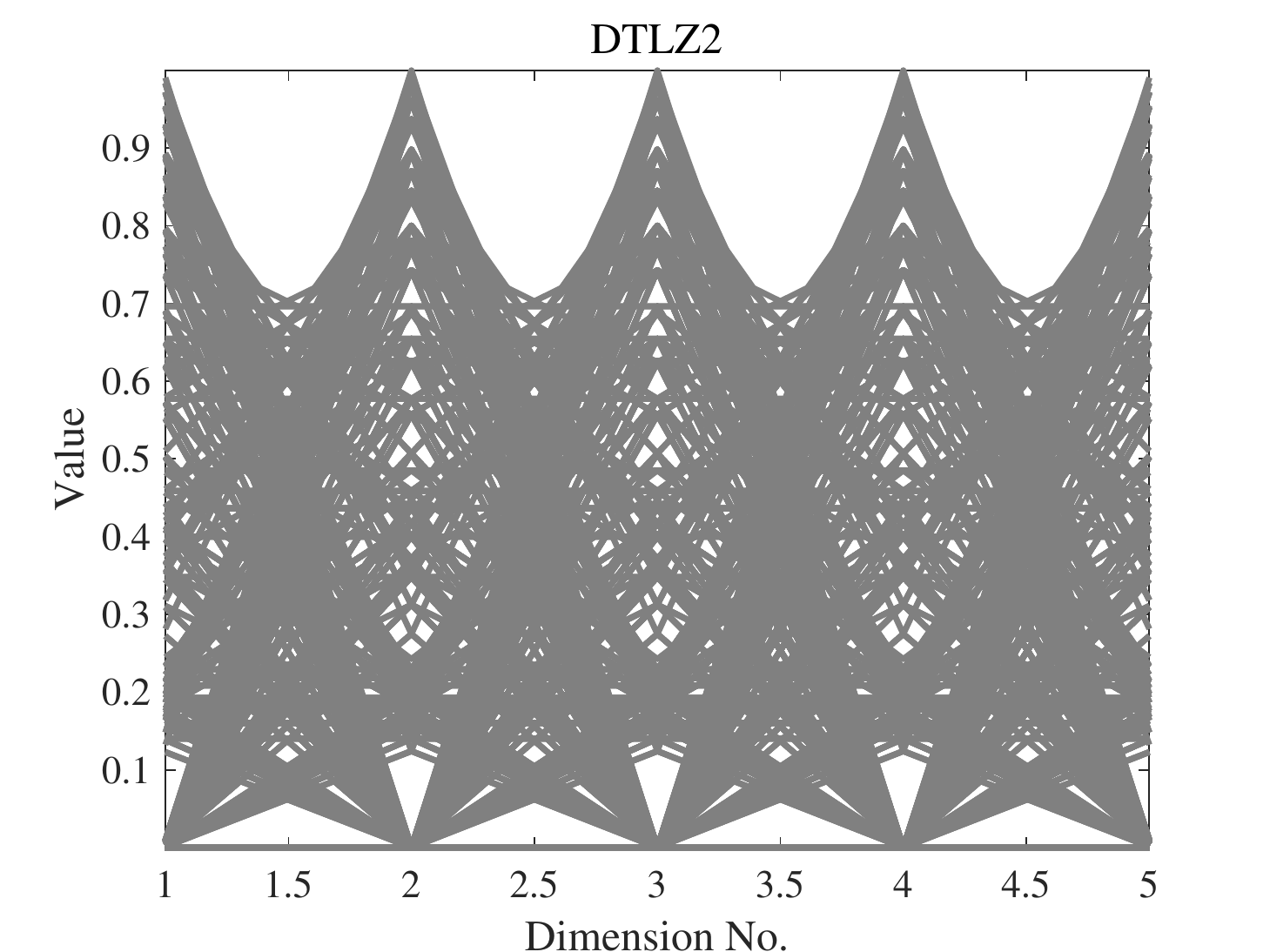}}

    \centerline{(f)  True PF}

  \end{minipage}
  \caption{Distributions of solutions on \mbox{DTLZ1-DTLZ2} with 5~objectives}
  \label{rand_pic}
\end{figure*}

We now compare the distribution of the offspring obtained by DPPs-Selection and uniform sampling.
We therefore run experiments on \mbox{DTLZ1-DTLZ2} with 5~objectives.
The population size is set to~126 and the number of function evaluations is~100'000.
We perform 30~runs per instance. DPPs-Selection provides a uniform distribution of solutions, as shown in Figure~\ref{rand_pic}. In contrast, the solutions from uniform sampling are clustered around several solutions.%
\subsection{Application Analysis}%
We now compare the CPU time of different algorithms and then investigate which types of problems MaOEADPPs is suitable for. We use instances of DTLZ1 to DTLZ8 with~5 and 10~objectives. The population size is set to~126 for~5 and to~230 for 10~objectives. The number of function evaluations is set to~100'000. We again perform 30~independent runs for each instance.

Table~\ref{run_time} shows the consumed CPU time (measured in seconds) of MaOEADPPs and the other four algorithms. We can see that RVEA is the fastest. MaOEADPPs needs more time than RVEA, GrEA, RSEA, and \mbox{NSGA-III}, but less than \mbox{AR-MOEA}. The time complexity of MaOEADPPs is higher, but based on the excellent results discussed in the previous sections, this seems acceptable to us. Most of the computing time is spent on the eigen-decomposition.
\begin{table*}[tbp]
  \renewcommand{\arraystretch}{1.2}
  \centering
  \caption{CPU TIME OF MAOEADPPS AND FOUR STATE-OF-THE-ART ALGORITHMS. THE BEST RESULTS FOR EACH INSTANCE ARE HIGHLIGHTED IN GRAY.}
  \resizebox{\textwidth}{!}{
    \begin{tabular}{cccccccc}
      \toprule
      Problem&$M$&$D$&AR-MOEA&\mbox{NSGA-III}&RVEA&GrEA&MaOEADPPs\\
      \midrule
      \multirow{1}{*}{DTLZ1}&5&9&6.5661e+1 (2.47e+0) $-$&3.8559e+0 (1.63e-1) $+$&\hl{2.7599e+0 (1.28e-1) $+$}&4.2230e+1 (1.01e+0) $+$&5.4494e+1 (4.52e+0)\\
      \multirow{1}{*}{DTLZ1}&10&14&3.3640e+2 (3.07e+1) $-$&5.8704e+0 (6.08e-1) $+$&\hl{2.7746e+0 (1.55e-1) $+$}&1.1005e+2 (3.26e+0) $+$&1.6037e+2 (1.18e+1)\\
      \hline
      \multirow{1}{*}{DTLZ2}&5&14&1.0239e+2 (5.01e+0) $-$&4.1235e+0 (1.70e-1) $+$&\hl{2.9035e+0 (1.76e-1) $+$}&5.5903e+1 (1.10e+0) $+$&6.0108e+1 (4.70e+0)\\
      \multirow{1}{*}{DTLZ2}&10&19&4.5566e+2 (4.97e+1) $-$&5.9594e+0 (1.04e+0) $+$&\hl{2.9470e+0 (2.92e-1) $+$}&1.1314e+2 (3.40e+0) $+$&2.0398e+2 (1.85e+1)\\
      \hline
      \multirow{1}{*}{DTLZ3}&5&14&5.6373e+1 (2.16e+0) $-$&4.0626e+0 (1.88e-1) $+$&\hl{2.9018e+0 (1.41e-1) $+$}&3.7598e+1 (1.14e+0) $+$&4.6408e+1 (3.66e+0)\\
      \multirow{1}{*}{DTLZ3}&10&19&3.2687e+2 (3.34e+1) $-$&6.1199e+0 (9.06e-1) $+$&\hl{2.8909e+0 (1.74e-1) $+$}&1.0895e+2 (2.69e+0) $+$&1.5727e+2 (1.50e+1)\\
      \hline
      \multirow{1}{*}{DTLZ4}&5&14&1.0739e+2 (6.12e+0) $-$&4.6813e+0 (8.36e-1) $+$&\hl{2.9570e+0 (2.25e-1) $+$}&5.7017e+1 (1.29e+0) $\approx$&6.0392e+1 (5.08e+0)\\
      \multirow{1}{*}{DTLZ4}&10&19&4.6751e+2 (4.96e+1) $-$&5.8799e+0 (1.05e+0) $+$&\hl{3.0433e+0 (1.68e-1) $+$}&1.1469e+2 (2.85e+0) $+$&2.0391e+2 (1.58e+1)\\
      \hline
      \multirow{1}{*}{DTLZ5}&5&14&1.0600e+2 (4.72e+0) $-$&6.2078e+0 (3.72e-1) $+$&\hl{2.3133e+0 (1.82e-1) $+$}&6.1369e+1 (1.36e+0) $+$&6.8480e+1 (6.22e+0)\\
      \multirow{1}{*}{DTLZ5}&10&19&4.3173e+2 (3.84e+1) $-$&7.6929e+0 (8.37e-1) $+$&\hl{2.3730e+0 (1.32e-1) $+$}&1.1278e+2 (2.89e+0) $+$&1.9501e+2 (1.44e+1)\\
      \hline
      \multirow{1}{*}{DTLZ6}&5&14&9.4896e+1 (4.01e+0) $-$&5.8468e+0 (3.78e-1) $+$&\hl{2.7999e+0 (2.08e-1) $+$}&5.6629e+1 (1.52e+0) $+$&6.1332e+1 (4.61e+0)\\
      \multirow{1}{*}{DTLZ6}&10&19&4.0879e+2 (3.60e+1) $-$&7.2795e+0 (9.04e-1) $+$&\hl{2.7355e+0 (1.78e-1) $+$}&1.1106e+2 (2.98e+0) $+$&1.6366e+2 (1.07e+1)\\
      \hline
      \multirow{1}{*}{DTLZ7}&5&24&1.0802e+2 (5.21e+0) $-$&6.7572e+0 (5.06e-1) $+$&\hl{2.4925e+0 (1.70e-1) $+$}&5.7722e+1 (1.42e+0) $+$&6.7323e+1 (4.35e+0)\\
      \multirow{1}{*}{DTLZ7}&10&29&4.9773e+2 (4.66e+1) $-$&7.4693e+0 (6.19e-1) $+$&\hl{2.3939e+0 (2.86e-1) $+$}&1.1469e+2 (3.21e+0) $+$&2.1071e+2 (1.67e+1)\\
      \hline
      \multirow{1}{*}{DTLZ8}&5&50&6.6489e+1 (4.64e+0) $-$&6.6228e+0 (7.32e-1) $+$&\hl{3.2762e+0 (2.31e-1) $+$}&4.0555e+1 (1.79e+0) $+$&4.2583e+1 (4.39e+0)\\
      \multirow{1}{*}{DTLZ8}&10&100&2.7658e+2 (1.47e+1) $-$&8.5522e+0 (8.12e-1) $+$&\hl{3.8377e+0 (2.41e-1) $+$}&1.2365e+2 (4.54e+0) $+$&1.8164e+2 (7.85e+0)\\
      \hline
      \bottomrule
  \end{tabular}}
  \label{run_time}
\end{table*}%
\begin{table*}[tbp]
  \renewcommand{\arraystretch}{1.2}
  \centering
  \caption{IGD VALUES OBTAINED BY MAOEADPPS, \mbox{AR-MOEA~\cite{tian2017indicator}}, \mbox{NSGA-III~\cite{deb2014evolutionary}}, RSEA~\cite{he2017radial}, \mbox{$\theta $-DEA~\cite{yuan2016new}}, AND RVEA~\cite{cheng2016reference} ON \mbox{DTLZ1-DTLZ4}, \mbox{IDTLZ1-IDTLZ2}, AND \mbox{WFG1-WFG9} WITH~20, 25 AND 30~OBJECTIVES. THE BEST RESULTS FOR EACH INSTANCE ARE HIGHLIGHTED IN GRAY.}
  \resizebox{\textwidth}{!}{
  \begin{tabular}{ccccccccc}
    \toprule
    Problem&$M$&$D$&MaOEADPPs&AR-MOEA&\mbox{NSGA-III}&RSEA&$\theta$-DEA&RVEA\\
    \midrule
    \multirow{1}{*}{DTLZ1}&20&24&\hl{1.6935e-1 (5.89e-3)}&1.8317e-1 (5.37e-3) $-$&2.0905e-1 (2.76e-2) $-$&2.5023e-1 (2.39e-2) $-$&2.4336e-1 (2.55e-2) $-$&1.7192e-1 (4.52e-3) $\approx$\\
    \multirow{1}{*}{DTLZ1}&25&29&\hl{1.5989e-1 (5.78e-3)}&2.0995e-1 (1.25e-2) $-$&3.4554e-1 (1.10e-1) $-$&2.5450e-1 (4.56e-2) $-$&3.0196e-1 (3.34e-2) $-$&1.8809e-1 (1.26e-2) $-$\\
    \multirow{1}{*}{DTLZ1}&30&34&\hl{1.7267e-1 (2.88e-3)}&2.0919e-1 (6.88e-3) $-$&2.8360e-1 (2.07e-2) $-$&2.2535e-1 (2.71e-2) $-$&3.1785e-1 (4.84e-2) $-$&1.9325e-1 (6.34e-3) $-$\\
    \hline
    \multirow{1}{*}{DTLZ2}&20&29&\hl{5.8814e-1 (1.44e-3)}&6.2045e-1 (1.18e-3) $-$&7.5759e-1 (4.03e-2) $-$&7.3204e-1 (2.25e-2) $-$&6.2287e-1 (2.65e-4) $-$&6.4054e-1 (3.92e-2) $-$\\
    \multirow{1}{*}{DTLZ2}&25&34&\hl{5.6679e-1 (1.08e-3)}&7.9821e-1 (2.37e-2) $-$&9.8159e-1 (2.44e-2) $-$&7.6549e-1 (2.13e-2) $-$&8.0077e-1 (2.64e-2) $-$&1.1378e+0 (1.33e-1) $-$\\
    \multirow{1}{*}{DTLZ2}&30&39&\hl{5.7159e-1 (2.15e-3)}&8.4406e-1 (2.54e-2) $-$&8.5350e-1 (9.35e-3) $-$&8.0201e-1 (3.75e-2) $-$&8.3091e-1 (1.29e-2) $-$&1.2173e+0 (1.48e-1) $-$\\
    \hline
    \multirow{1}{*}{DTLZ3}&20&29&6.3938e-1 (2.58e-2)&\hl{6.3548e-1 (2.40e-2) $\approx$}&5.8748e+1 (3.19e+1) $-$&2.0867e+0 (1.36e+0) $-$&8.7577e-1 (3.31e-1) $-$&6.4834e-1 (1.99e-2) $\approx$\\
    \multirow{1}{*}{DTLZ3}&25&34&\hl{6.9541e-1 (5.33e-2)}&9.1638e-1 (9.02e-2) $-$&4.4350e+0 (2.58e+0) $-$&3.4304e+0 (1.90e+0) $-$&1.3322e+0 (5.65e-1) $-$&1.1717e+0 (1.88e-1) $-$\\
    \multirow{1}{*}{DTLZ3}&30&39&\hl{9.1521e-1 (5.59e-2)}&9.4735e-1 (4.23e-2) $\approx$&1.9279e+0 (1.37e+0) $-$&4.7211e+0 (2.99e+0) $-$&1.2745e+0 (3.52e-1) $-$&1.1657e+0 (1.22e-1) $-$\\
    \hline
    \multirow{1}{*}{DTLZ4}&20&29&\hl{5.8886e-1 (1.26e-3)}&6.1904e-1 (6.32e-4) $-$&6.8387e-1 (5.20e-2) $-$&7.1766e-1 (1.34e-2) $-$&6.2301e-1 (5.88e-5) $-$&6.2318e-1 (7.11e-5) $-$\\
    \multirow{1}{*}{DTLZ4}&25&34&\hl{5.7182e-1 (4.78e-4)}&7.9480e-1 (5.26e-3) $-$&9.6287e-1 (3.85e-2) $-$&7.6557e-1 (1.67e-2) $-$&7.7492e-1 (1.29e-3) $-$&8.2984e-1 (2.57e-2) $-$\\
    \multirow{1}{*}{DTLZ4}&30&39&\hl{5.8104e-1 (5.87e-4)}&8.1673e-1 (3.02e-3) $-$&8.5469e-1 (1.01e-2) $-$&8.1422e-1 (1.09e-2) $-$&8.0093e-1 (1.30e-3) $-$&8.2006e-1 (2.23e-3) $-$\\
    \hline
    \multirow{1}{*}{IDTLZ1}&20&24&\hl{1.7950e-1 (3.48e-3)}&2.0867e-1 (1.70e-2) $-$&2.2502e-1 (1.07e-2) $-$&2.1841e-1 (1.00e-2) $-$&2.3480e-1 (8.83e-3) $-$&3.9018e-1 (3.56e-2) $-$\\
    \multirow{1}{*}{IDTLZ1}&25&29&\hl{1.7689e-1 (6.70e-3)}&2.5662e-1 (1.58e-2) $-$&2.2169e-1 (4.88e-3) $-$&2.1581e-1 (8.49e-3) $-$&2.1597e-1 (5.56e-3) $-$&3.6129e-1 (4.53e-2) $-$\\
    \multirow{1}{*}{IDTLZ1}&30&34&\hl{1.8500e-1 (7.06e-3)}&2.5841e-1 (1.41e-2) $-$&2.2867e-1 (4.67e-3) $-$&2.2556e-1 (8.28e-3) $-$&2.1883e-1 (5.43e-3) $-$&4.0375e-1 (4.15e-2) $-$\\
    \hline
    \multirow{1}{*}{IDTLZ2}&20&29&\hl{6.0817e-1 (2.84e-3)}&7.5260e-1 (1.40e-2) $-$&8.2186e-1 (1.47e-2) $-$&8.9332e-1 (8.68e-3) $-$&1.0703e+0 (4.38e-3) $-$&1.0573e+0 (6.83e-3) $-$\\
    \multirow{1}{*}{IDTLZ2}&25&34&\hl{5.6219e-1 (2.99e-3)}&8.8987e-1 (3.19e-2) $-$&9.4107e-1 (3.26e-2) $-$&8.8319e-1 (8.75e-3) $-$&1.0284e+0 (9.56e-3) $-$&1.1280e+0 (1.61e-2) $-$\\
    \multirow{1}{*}{IDTLZ2}&30&39&\hl{5.7035e-1 (2.15e-3)}&9.4525e-1 (4.04e-2) $-$&9.5061e-1 (6.20e-2) $-$&9.3419e-1 (3.32e-3) $-$&9.8636e-1 (8.10e-2) $-$&1.1657e+0 (5.73e-3) $-$\\
    \hline
    \multirow{1}{*}{WFG1}&20&29&\hl{2.8416e+0 (5.71e-2)}&4.3317e+0 (1.07e-1) $-$&4.5705e+0 (2.50e-1) $-$&3.3638e+0 (7.35e-2) $-$&3.8257e+0 (1.55e-1) $-$&3.9325e+0 (8.88e-2) $-$\\
    \multirow{1}{*}{WFG1}&25&34&\hl{2.6685e+0 (1.38e-1)}&4.3395e+0 (4.52e-2) $-$&4.6421e+0 (1.74e-1) $-$&3.2426e+0 (1.57e-1) $-$&4.1055e+0 (1.24e-1) $-$&4.2723e+0 (7.51e-2) $-$\\
    \multirow{1}{*}{WFG1}&30&39&\hl{3.1385e+0 (6.05e-2)}&5.0391e+0 (7.49e-2) $-$&5.0819e+0 (3.44e-1) $-$&4.0689e+0 (2.29e-1) $-$&4.8393e+0 (1.43e-1) $-$&4.8916e+0 (9.55e-2) $-$\\
    \hline
    \multirow{1}{*}{WFG2}&20&29&\hl{2.7857e+0 (5.94e-2)}&3.3199e+0 (1.81e-1) $-$&4.0051e+0 (1.70e-1) $-$&4.2709e+0 (4.71e-1) $-$&8.3963e+0 (2.41e-1) $-$&3.3546e+0 (1.54e-1) $-$\\
    \multirow{1}{*}{WFG2}&25&34&\hl{2.5679e+0 (9.42e-2)}&3.4314e+0 (1.50e-1) $-$&7.7962e+0 (2.91e+0) $-$&5.5383e+0 (1.36e+0) $-$&1.4250e+1 (2.86e+0) $-$&4.2759e+0 (8.53e-2) $-$\\
    \multirow{1}{*}{WFG2}&30&39&\hl{2.9826e+0 (8.56e-2)}&4.1345e+0 (1.51e-1) $-$&4.7334e+0 (4.81e-1) $-$&6.4618e+0 (1.45e+0) $-$&1.7321e+1 (4.11e+0) $-$&4.7467e+0 (1.37e-1) $-$\\
    \hline
    \multirow{1}{*}{WFG3}&20&29&3.8937e+0 (3.58e-1)&7.6670e+0 (2.35e-1) $-$&8.9111e+0 (2.21e+0) $-$&3.6430e+0 (2.10e+0) $\approx$&\hl{2.8643e+0 (2.48e-1) $+$}&1.0329e+1 (3.05e+0) $-$\\
    \multirow{1}{*}{WFG3}&25&34&\hl{5.0618e+0 (7.49e-1)}&9.9951e+0 (1.63e+0) $-$&1.8565e+1 (4.36e+0) $-$&7.1375e+0 (3.62e+0) $-$&8.0814e+0 (9.90e-1) $-$&1.5028e+1 (2.84e+0) $-$\\
    \multirow{1}{*}{WFG3}&30&39&4.5820e+0 (7.64e-1)&1.2189e+1 (1.97e+0) $-$&8.9821e+0 (1.58e+0) $-$&\hl{4.1652e+0 (2.67e+0) $\approx$}&9.5068e+0 (1.14e+0) $-$&1.7722e+1 (1.14e+0) $-$\\
    \hline
    \multirow{1}{*}{WFG4}&20&29&\hl{1.0771e+1 (1.46e-1)}&1.1501e+1 (4.03e-2) $-$&1.2316e+1 (6.83e-1) $-$&1.3459e+1 (3.95e-1) $-$&1.1459e+1 (1.26e-2) $-$&1.2430e+1 (3.98e-1) $-$\\
    \multirow{1}{*}{WFG4}&25&34&\hl{1.1448e+1 (7.12e-2)}&2.2522e+1 (1.43e+0) $-$&2.0519e+1 (8.80e-1) $-$&1.8238e+1 (5.28e-1) $-$&2.1231e+1 (2.50e-1) $-$&2.1083e+1 (9.00e-1) $-$\\
    \multirow{1}{*}{WFG4}&30&39&\hl{1.4064e+1 (3.89e-1)}&2.9242e+1 (1.40e+0) $-$&2.5464e+1 (2.70e-1) $-$&2.2991e+1 (3.21e-1) $-$&2.6360e+1 (2.48e-1) $-$&2.5982e+1 (2.98e-1) $-$\\
    \hline
    \multirow{1}{*}{WFG5}&20&29&\hl{1.0807e+1 (3.45e-2)}&1.1498e+1 (2.51e-2) $-$&1.1919e+1 (8.91e-1) $-$&1.3100e+1 (2.55e-1) $-$&1.1311e+1 (6.55e-2) $-$&1.1781e+1 (9.05e-2) $-$\\
    \multirow{1}{*}{WFG5}&25&34&\hl{1.1298e+1 (3.73e-2)}&2.0772e+1 (7.94e-1) $-$&2.0517e+1 (6.45e-1) $-$&1.6908e+1 (3.24e-1) $-$&2.0838e+1 (3.52e-1) $-$&2.1135e+1 (1.02e+0) $-$\\
    \multirow{1}{*}{WFG5}&30&39&\hl{1.3500e+1 (1.68e-1)}&2.7843e+1 (2.48e+0) $-$&2.5424e+1 (1.85e-1) $-$&2.1837e+1 (6.32e-1) $-$&2.5355e+1 (2.20e-1) $-$&2.6277e+1 (6.84e-1) $-$\\
    \hline
    \multirow{1}{*}{WFG6}&20&29&\hl{1.0545e+1 (2.04e-1)}&1.1564e+1 (2.40e-2) $-$&1.4251e+1 (7.35e-1) $-$&1.4219e+1 (3.13e-1) $-$&1.1519e+1 (1.77e-2) $-$&1.3099e+1 (2.28e-1) $-$\\
    \multirow{1}{*}{WFG6}&25&34&\hl{1.1426e+1 (6.55e-2)}&2.2205e+1 (1.01e+0) $-$&2.1670e+1 (9.75e-1) $-$&1.9284e+1 (9.24e-1) $-$&2.1142e+1 (2.76e-1) $-$&2.4464e+1 (2.98e+0) $-$\\
    \multirow{1}{*}{WFG6}&30&39&\hl{1.3568e+1 (4.89e-2)}&2.7633e+1 (1.42e+0) $-$&2.5705e+1 (7.05e-1) $-$&2.4238e+1 (1.02e+0) $-$&2.5888e+1 (3.09e-1) $-$&3.0529e+1 (1.93e+0) $-$\\
    \hline
    \multirow{1}{*}{WFG7}&20&29&\hl{1.0701e+1 (5.17e-2)}&1.1652e+1 (4.25e-2) $-$&1.4793e+1 (9.31e-1) $-$&1.4335e+1 (5.76e-1) $-$&1.1494e+1 (4.67e-2) $-$&1.1831e+1 (3.53e-1) $-$\\
    \multirow{1}{*}{WFG7}&25&34&\hl{1.1613e+1 (8.21e-2)}&2.2584e+1 (7.95e-1) $-$&2.0911e+1 (1.06e+0) $-$&1.9592e+1 (1.15e+0) $-$&2.1747e+1 (7.73e-1) $-$&2.7103e+1 (1.06e+1) $-$\\
    \multirow{1}{*}{WFG7}&30&39&\hl{1.4747e+1 (2.94e-1)}&2.9547e+1 (9.79e-1) $-$&2.5760e+1 (6.03e-2) $-$&2.5355e+1 (1.04e+0) $-$&2.6840e+1 (6.37e-1) $-$&2.7464e+1 (1.68e+0) $-$\\
    \hline
    \multirow{1}{*}{WFG8}&20&29&\hl{1.0602e+1 (1.27e-1)}&1.1549e+1 (2.23e-2) $-$&1.3472e+1 (1.37e+0) $-$&1.6935e+1 (4.24e-1) $-$&1.2782e+1 (3.53e-1) $-$&1.2662e+1 (7.01e-1) $-$\\
    \multirow{1}{*}{WFG8}&25&34&\hl{1.3453e+1 (1.61e+0)}&2.3285e+1 (1.02e+0) $-$&2.2056e+1 (1.86e+0) $-$&2.2662e+1 (6.99e-1) $-$&2.1532e+1 (6.84e-1) $-$&2.6385e+1 (7.50e+0) $-$\\
    \multirow{1}{*}{WFG8}&30&39&\hl{1.6808e+1 (2.42e+0)}&3.1358e+1 (8.10e-1) $-$&2.6514e+1 (2.87e-1) $-$&2.9600e+1 (8.20e-1) $-$&2.7165e+1 (5.15e-1) $-$&2.9285e+1 (3.28e+0) $-$\\
    \hline
    \multirow{1}{*}{WFG9}&20&29&\hl{1.0970e+1 (1.34e-1)}&1.1725e+1 (1.07e-1) $-$&1.3277e+1 (1.49e+0) $-$&1.3659e+1 (3.04e-1) $-$&1.1386e+1 (2.58e-1) $-$&1.1674e+1 (3.53e-1) $-$\\
    \multirow{1}{*}{WFG9}&25&34&\hl{1.2027e+1 (5.54e-1)}&2.2138e+1 (6.15e-1) $-$&2.0130e+1 (4.52e-1) $-$&1.7073e+1 (3.48e-1) $-$&2.0740e+1 (2.10e-1) $-$&2.1086e+1 (1.52e+0) $-$\\
    \multirow{1}{*}{WFG9}&30&39&\hl{1.5320e+1 (7.84e-1)}&2.9121e+1 (9.59e-1) $-$&2.5257e+1 (7.76e-1) $-$&2.1094e+1 (6.04e-1) $-$&2.5459e+1 (3.21e-1) $-$&2.5826e+1 (2.89e+0) $-$\\
    \multicolumn{4}{c}{$+/-/\approx$}&0/43/2&0/45/0&0/43/2&1/44/0&0/43/2\\
    \bottomrule
  \end{tabular}}
  \label{large_objective}
\end{table*}

We believe that MaOEADPPs is suitable for solving problems which are not time critical but sufficiently difficult.
From another perspective, it is also suitable for problems where the objective function evaluations are time consuming, because in this case, the runtime requirements of MaOEADPPs will not play an important role.
For example, many application scenarios such as engineering design optimization, damage detection and pipeline systems~\cite{MahdaviMetaheuristics} have \textcolor{blue}{a large number of} decision variables.
These applications usually do not require very quick responses but improvements of solution quality can have \textcolor{blue}{a positive impact.}

We thus compare our MaOEADPPs with the other algorithms on the 8~large-scale instances LSMOP1 to LSMOP5, MONRP, SMOP1 and MOKP with~5 or 2~objectives.
We again perform 30~independent runs per instance.
The population size is~126 and the number of function evaluations is set to~100'000.
As shown in Table~\ref{decsion}, MaOEADPPs obtains the best IGD values on 6~of the 8~instances.

MaOEADPPs is also suitable for problems with a large number of objectives. To verify this, we carry out comparison experiments on DTLZ1-DTLZ4, WFG1-WFG9 and IDTLZ1-IDTLZ2 with~20, 25 and 30~objectives.
We perform again 30~independent runs per instance.
We set the population size to~400 for instances with 30~objectives and to~300 for instances with~20 and 25~objectives.
The number of function evaluations is set to~100'000.

As shown in Table~\ref{large_objective}, our MaOEADPPs returns the best results on~42 of the~45 instances.
Moreover, as the number of objectives increases, the gaps between the IGD values of our MaOEADPPs and the compared algorithms gradually widen.
One possible reason is that MaOEADPPs uses the Kernel Matrix~$\boldsymbol{L}$ to measure the convergence and similarity between solution pairs, which reflects more details of the distribution in the population.%
\begin{table*}[tbp]
  \renewcommand{\arraystretch}{1.2}
  \centering
  \caption{IGD VALUES OF \mbox{AR-MOEA~\cite{tian2017indicator}}, \mbox{NSGA-III~\cite{deb2014evolutionary}}, RSEA~\cite{he2017radial} AND \mbox{$\theta $-DEA~\cite{yuan2016new}} ON PROBLEMS WITH A LARGE NUMBER OF DECISION VARIABLES. THE BEST RESULTS FOR EACH INSTANCE ARE HIGHLIGHTED IN GRAY.}
  \resizebox{\textwidth}{!}{
  \begin{tabular}{cccccccc}
    \toprule
    Problem&$M$&$D$&MaOEADPPs&AR-MOEA&\mbox{NSGA-III}&$\theta-$DEA&RSEA\\
    \midrule
    \multirow{1}{*}{SMOP1}&5&100&\hl{1.2944e-1 (1.30e-3)}&1.3449e-1 (1.69e-3) $-$&1.3601e-1 (2.19e-3) $-$&1.3093e-1 (1.66e-3) $-$&1.5153e-1 (3.75e-3) $-$\\
    \hline
    \multirow{1}{*}{MOKP}&5&250&1.7386e+4 (2.25e+2)&1.7528e+4 (9.28e+1) $-$&\hl{1.7168e+4 (1.62e+2) $+$}&1.7337e+4 (1.12e+2) $\approx$&1.7526e+4 (1.12e+2) $-$\\
    \hline
    \multirow{1}{*}{MONRP}&2&100&\hl{2.7285e+3 (5.82e+2)}&8.9542e+3 (1.26e+3) $-$&7.6479e+3 (7.22e+2) $-$&8.4295e+3 (9.67e+2) $-$&9.9819e+3 (1.08e+3) $-$\\
    \hline
    \multirow{1}{*}{LSMOP1}&5&500&1.4587e+0 (1.97e-1)&3.8753e+0 (4.51e-1) $-$&3.3043e+0 (4.01e-1) $-$&9.0859e-1 (8.97e-2) $+$&\hl{7.8286e-1 (1.10e-1) $+$}\\
    \hline
    \multirow{1}{*}{LSMOP2}&5&500&\hl{1.6795e-1 (1.03e-3)}&1.7401e-1 (4.16e-4) $-$&1.7492e-1 (7.40e-4) $-$&1.7411e-1 (6.70e-4) $-$&1.8579e-1 (3.41e-3) $-$\\
    \hline
    \multirow{1}{*}{LSMOP3}&5&500&\hl{5.2013e+0 (7.16e-1)}&7.7309e+0 (1.08e+0) $-$&9.5593e+0 (1.42e+0) $-$&9.8999e+0 (3.24e+0) $-$&9.2354e+0 (2.28e+0) $-$\\
    \hline
    \multirow{1}{*}{LSMOP4}&5&500&\hl{2.8271e-1 (5.33e-3)}&3.0373e-1 (4.04e-3) $-$&3.0943e-1 (5.85e-3) $-$&2.9601e-1 (4.46e-3) $-$&2.9578e-1 (1.52e-2) $-$\\
    \hline
    \multirow{1}{*}{LSMOP5}&5&500&\hl{2.1142e+0 (7.12e-1)}&3.4117e+0 (9.69e-1) $-$&6.9847e+0 (7.16e-1) $-$&4.7954e+0 (5.67e-1) $-$&3.6352e+0 (6.41e-1) $-$\\
    \hline
    \multicolumn{4}{c}{$+/-/\approx$}&0/8/0&1/7/0&1/6/1&1/7/0\\
    \bottomrule
  \end{tabular}}
  \label{decsion}
\end{table*}%
\section{CONCLUSION}%
\label{sec:conclusios}%
In this paper, we propose a new MaOEA named MaOEADPPs. We show that the algorithm is suitable for solving a variety of types of MaOPs. For MaOEADPPs, we design the new Determinantal Point Processes Selection (DPPs-Selection) method for the Environmental Selection. In order to adapt to various types of MaOPs, a Kernel Matrix is defined. We compare MaOEADPPs with several state-of-the-art algorithms on the DTLZ, WFG, and MaF benchmarks with the number of objectives ranging from~5 to~30. The experimental results demonstrate that our MaOEADPPs significantly outperforms the other algorithms. We find that the Kernel Matrix of MaOEADPPs can adapt to problems with different objective types. Our results also suggest that MaOEADPPs is not the fastest algorithm but also not slow. It is therefore suitable for solving problems where the objective function evaluations are time consuming or which are not time critical. The Kernel Matrix of the DPPs-Selection is the key component of our MaOEADPPs. We will try to further improve it in the future work and especially focus on investigating alternative definitions for the solutions quality metrics.%
\section*{ACKNOWLEDGEMENT}
This research work was supported by the National Natural Science Foundation of China under Grant~61573328.%
%
\bibliography{new_paper}%
\bibliographystyle{IEEEtran}%
\end{document}